\documentclass[12pt]{article}
\usepackage[letterpaper, scale=0.8]{geometry}
\usepackage{amsmath}
\usepackage{amsfonts}
\usepackage{physics}
\usepackage{graphicx}
\usepackage{float}
\usepackage{booktabs}
\usepackage{multirow}

\usepackage{algpseudocode}
\usepackage[ruled, vlined, linesnumbered]{algorithm2e}

\usepackage{cite}

\usepackage{amsthm}
\usepackage{xcolor}

\usepackage{amsfonts}
\usepackage[labelfont=bf]{caption}
\usepackage{float}

\usepackage{authblk}
\usepackage{hyperref}
\usepackage{graphicx}
\usepackage{placeins}

\usepackage{wrapfig}
\usepackage{url}            
\usepackage{booktabs}       
\usepackage{amsfonts}       
\usepackage{nicefrac}       
\usepackage{microtype}      
\usepackage{xcolor}         
\usepackage[caption = false]{subfig}

\usepackage[skip=8pt, indent=18pt]{parskip}        

\usepackage[utf8]{inputenc} 
\usepackage[T1]{fontenc}    
\usepackage{hyperref}       
\usepackage{url}            
\usepackage{booktabs}       
\usepackage{amsfonts}       
\usepackage{nicefrac}       
\usepackage{microtype}      
\usepackage{xcolor}         

\usepackage{amsfonts, amsmath,amssymb,array,latexsym, hyperref, amscd}
\usepackage{microtype}
\usepackage{multirow}
\usepackage{comment}

\usepackage{fancyhdr,a4wide}
\usepackage{amsthm}
\usepackage{graphicx}
\usepackage[dvipsnames]{xcolor}
\usepackage{color}
\usepackage{enumitem}

\usepackage{subcaption}

\usepackage{xr-hyper} 
\numberwithin{equation}{section}

\newcommand{\set}[1]{\left\{#1\right\}}

\newcommand{\gives}{\rightarrow}

\newcommand{\mcL}{\mathcal L}

\hypersetup{colorlinks=true,allcolors=[rgb]{0.0,0.4,0.6}}

\usepackage[noabbrev]{cleveref}
\Crefformat{figure}{Fig.~(#2#1#3)}
\crefformat{figure}{fig.~(#2#1#3)}
\Crefformat{section}{Section~(#2#1#3)}
\crefformat{section}{section~(#2#1#3)}
\Crefformat{equation}{Eq.~(#2#1#3)}
\crefformat{equation}{eq.~(#2#1#3)}
\Crefformat{table}{Table~(#2#1#3)}
\crefformat{table}{table~(#2#1#3)}
\Crefformat{appendix}{Appendix~(#2#1#3)}
\crefformat{appendix}{appendix~(#2#1#3)}
\Crefformat{algorithm}{Algorithm~(#2#1#3)}
\crefformat{algorithm}{algorithm~(#2#1#3)}

\usepackage{amsmath,amsfonts,bm}

\def\1{\bm{1}}

\DeclareMathAlphabet{\mathsfit}{\encodingdefault}{\sfdefault}{m}{sl}
\SetMathAlphabet{\mathsfit}{bold}{\encodingdefault}{\sfdefault}{bx}{n}

\newcommand{\E}{\mathbb{E}}

\newcommand{\R}{\mathbb{R}}

\newcommand{\Var}{\mathrm{Var}}

\makeatletter
\renewenvironment{abstract}{%
    \if@twocolumn
      \section*{\abstractname}%
    \else 
      \begin{center}%
        {\bfseries \large\abstractname\vspace{\z@}}
      \end{center}%
      \quotation
    \fi}
    {\if@twocolumn\else\endquotation\fi}
\makeatother

\theoremstyle{plain}
\theoremstyle{definition}
\theoremstyle{remark}

\algrenewcommand\algorithmicensure{\textbf{Output:}}

\definecolor{grey}{rgb}{0.4, 0.4, 0.4}    
\definecolor{olive}{rgb}{0.5, 0.5, 0}   

\title{Hyperparameter Transfer for Dense Associative Memories}

\author[1]{Roi Holtzman}
\author[2]{Dmitry Krotov}
\author[3]{Boris Hanin}

\affil[1]{\small Rudolf Peierls Centre for Theoretical Physics, University of Oxford,  
  Oxford OX1 3PU, UK }
\affil[2]{Dynamical Mind, IBM Research }
\affil[3]{Princeton ORFE }
\date{}

\begin{document}
\maketitle

\begin{abstract}
Dense Associative Memory (DenseAM) is a promising family of AI architectures that is represented by a neural network performing temporal dynamics on an energy landscape. While hyperparameter transfer methods are well-studied for feed-forward networks, these methods have not been developed for settings in which weights are shared across layers and within the layer, which is common in DenseAMs. Additionally, DenseAMs utilize rapidly peaking activation functions that are rarely used in feed-forward architectures. The confluence of these aspects makes DenseAM a challenging framework for using existing methods for hyperparameter transfer. Our work initiates the development of hyperparameter transfer methods for this class of models. We derive explicit prescriptions for how the hyperparameters tuned on small models can be transferred to models trained at scale. We demonstrate excellent agreement between these theoretical findings and empirical results. 
\end{abstract}

\newpage

\section{Introduction}
The empirical success of modern deep learning has been driven by dramatic increases in model size, dataset size, and compute budget. To realize the performance gains possible at large scale, for every group of trainable parameters practitioners must carefully tune a set of hyperparameters (HPs), such as initialization scale and learning rate. Since such tuning is prohibitively expensive when done directly at large scale it is common instead to aim for \textit{hyperparameter transfer}, in which one first finds empirically good settings of HPs on small models and datasets and uses them to estimate performant HPs at larger scale. 

Theoretically guided approaches to HP transfer, often built on the $\mu$P heuristic  \cite{yang2020feature} and its variants \cite{everett2024scalingexponentsparameterizationsoptimizers, Dey25Dont, yang2023spectral}, have now been developed for neural architectures ranging from MLPs \cite{yang2020feature} to  dense transformers \cite{bordelon2024infinite,Dey25Dont} and sparse transformers with MoE layers \cite{jiang2026hyperparameter}. The core idea in these papers is to perform a dimensional analysis to determine how to scale HPs as functions of growing model size in such a way that training dynamics are consistent across scale and well-posed in the limit of infinite scale. Models and datasets at any finite scale are then discretizations of the same idealized infinite size limit, and one identifies \textit{dimensionless} versions of HPs that can be tuned at small scale.

In our paper this strategy is adapted for Dense Associative Memories (DenseAMs), which is a popular class of AI models whose forward pass describes the temporal evolution of a state vector on an energy landscape \cite{krotov2016dense}. As we explain in Sec. \ref{sec:DAM}, DenseAMs are energy-based associative memory networks \cite{krotov2025modern}. They generalize the celebrated Hopfield networks \cite{hopfield1982neural}, which are toy model systems performing memory retrieval through dynamical evolution of a network state. Despite their mathematical elegance, Hopfield networks are known to have a very small information storage capacity \cite{amit1985storing}, which limits their practical use in large scale AI systems \cite{krotov2023new}. DenseAMs have superior scaling laws for the amount of information that can be stored as a function of the network size \cite{krotov2016dense}. They operate in the same configuration space as Hopfield networks, but permit a much denser packing of memories. Although initially designed for efficient information storage and retrieval, modern use cases of DenseAMs extend far beyond a simple memory retrieval task and enable the  design of general compute architectures.

For instance, Energy Transformers \cite{hoover2023energy} evolve a collection of tokens on a learned energy landscape, dynamically transforming an initial masked state to a final configuration that can be decoded into meaningful auto-completions of images, graphs, and even solutions to PDEs \cite{zhang2025operator}. Energy-based GPT models (NRGPT) use dynamical evolution of tokens on the energy landscape to predict the next token in a sentence \cite{dehmamy2025nrgpt}. In all these examples DenseAMs are used as sub-modules of more general architectures that evolve the state of the network (e.g., a collection of tokens) according to non-linear dynamical equations describing energy decent dynamics. The computational program performed by these systems is encoded in the shape of this energy landscape.    

\begin{figure}
    \centering
    \includegraphics[width=1\linewidth]{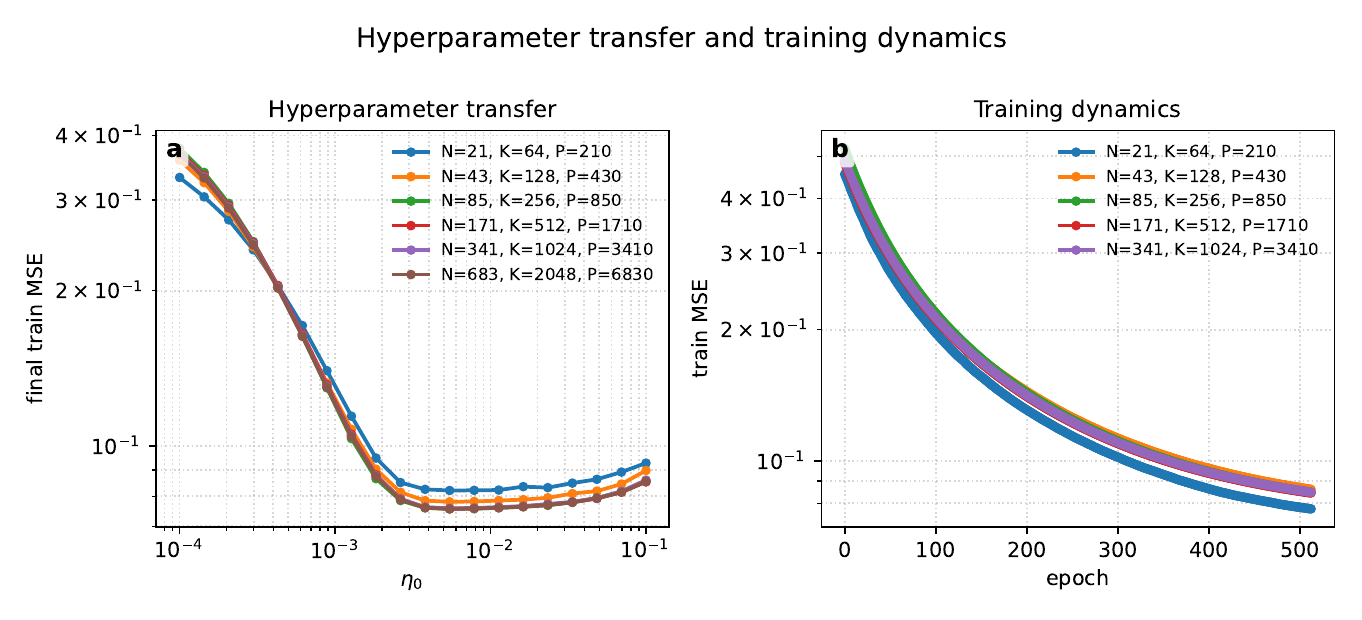}
    \caption{
Learning rate transfer and training loss dynamics for a linear DenseAM trained on a denoising objective \eqref{eq:loss} using mini-batch SGD in the proportional regime \eqref{eq:proportional-regime} with scale factors $\kappa=3, \rho=10, \beta=0.1$ and $256$ epochs. Here $\eta_0$ is the \textit{effective learning rate} (see Table \ref{tab:results}).
    }
    \label{fig:linear_SGD}
\end{figure}

Despite increasing significance of these ideas in the general landscape of AI models, at present it is unknown how to perform the HP transfer in these settings. This is not an easy task. DenseAM family differs from the conventional feed-forward settings in a number of aspects. For instance, the existence of an energy function requires sharing weights across multiple layers and operations within the layer. Additionally, dynamical equations describing energy-descent dynamics natively contain residual connections. Lastly, the shape of the energy landscape, which leads to superior scaling laws for the information storage, requires the use of steep activations functions, which are rarely used in more common feed-forward settings. The confluence of these aspects makes it difficult to apply results from known HP transfer methods in feed-forward networks to these new problems.  

The goal of our present paper is to initiate the study of $\mu$P-type HP transfer for the simplest DenseAM setting, which retains core aspects of network dynamics common in DenseAM family. Our network is described by the following equation 
\begin{equation}
\label{eq:f-def}
    f(x) =f_W(x)= s_2 W^\top\sigma\big[s_1 W g(x) + b\big]+ c,\quad x,g(x),c\in \R^N, \, b \in \R^K, \, W\in \R^{K\times N}, 
\end{equation}
which is an update rule that performs a single step in the direction of (negative) gradient of the energy. The construction of the energy landscape and network dynamics is explained in Sec. \ref{sec:DAM}. In \eqref{eq:f-def}, the scalars $s_1,s_2$ determine initialization scale, $g:\R\gives \R$ is a fixed non-linearity applied pointwise to $x$, and $\sigma$ may be either a neuron-wise non-linearity or a contrastive non-linearity, such as softmax.  At training time, we focus on minimizing a de-noising loss function
\begin{equation}
\label{eq:loss}
\mathcal L(W):=\frac{1}{2B}\sum_{\mu=1}^B \norm{f_W(x^\mu + \epsilon^\mu) - x^\mu}^2,\qquad \epsilon^\mu \sim \mathcal{N}(0, \sigma_\epsilon^2 I_{N\times N})
\end{equation}
using SGD or Adam, where $\left\{x^\mu\right\}_{\mu=1}^B$ are the datapoints in the batch. Our goal is to predict how near-optimal settings of scales $s_1,s_2$ and weight/bias learning rates $\eta_W,\eta_b,\eta_c$ scale with input/output dimension $N$, hidden layer width $K$, dataset size $P$, and batch size $B$. Our contributions are:

\begin{figure}
    \centering
    \includegraphics[width=1.0\linewidth]{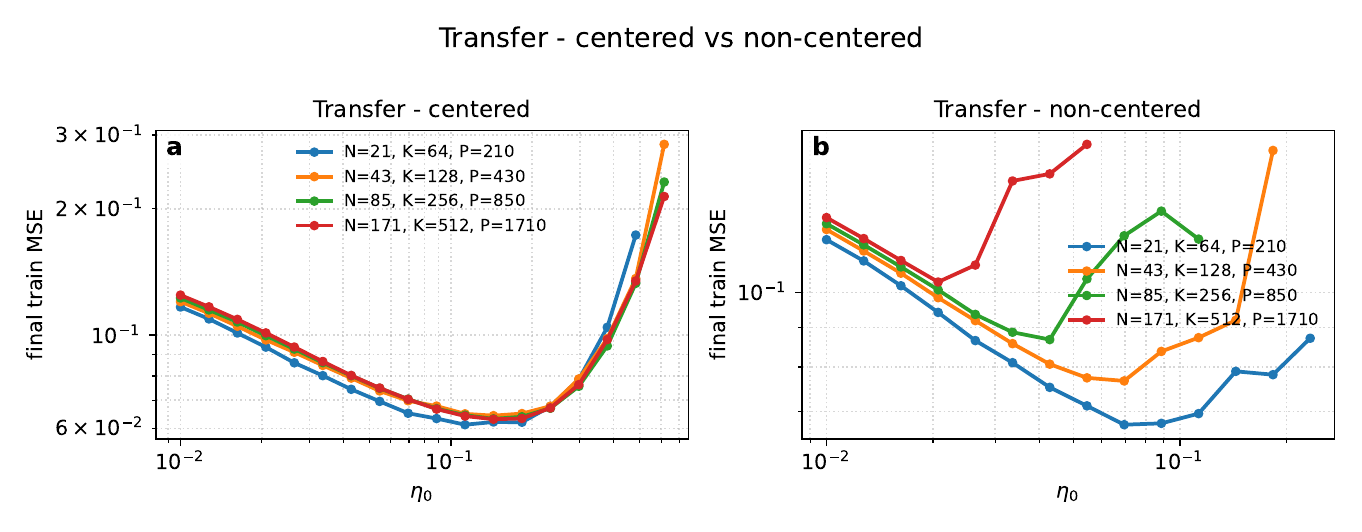}
    \caption{
    Learning rate transfer for a ReLU DenseAM trained by minimizing the denoising objective \eqref{eq:loss} using mini-batch SGD in the proportional regime \eqref{eq:proportional-regime} with scale factors $\kappa=3, \rho=10, \beta=0.1$ and $256$ epochs. Here $\eta_0$ is the \textit{effective learning rate} from \eqref{eq:sgd-linear-hp} and Table \ref{tab:results}. Non-linearity centering is critical for training stability and therefore transfer.
    }
    \label{fig:relu1-center-non-center}
\end{figure}

\begin{itemize}
    \item We provide a theoretical analysis of HP transfer when simultaneously scaling $N,K,P,B$ in the fundamental setting of linear activation $\sigma$ trained on isotropic data $x^\mu\sim \mathcal{N}(0,I_N)$. In this context, we find that all internal statistics of model training dynamics, including training loss and hidden layer feature covariances, collapse in the proportional scaling regime 
    \begin{equation}\label{eq:proportional-regime}
        N\gives \infty, \quad K = \kappa * N, \quad P = \rho * N, \quad B = \beta * P,\qquad \kappa, \rho, \beta \text{ fixed}.
    \end{equation}
    See Figure \ref{fig:linear_SGD}. Consistency across scales requires changing $s_1, s_2$ and learning rates as functions of $N$ as prescribed in Table \ref{tab:results}. See Sec. \ref{sec:empirics} for more empirical results on linear DenseAMs, Sec. \ref{sec:approach} for an overview of how we analyze DenseAMs in the proportional regime, and the Appendix for full derivations.

    \item For non-linear activations $\sigma$, a simple computation suggests that stable training in DenseAMs \eqref{eq:f-def} is possible only after removing the layer-wise mean of \textit{both pre and post-activations}. See Figures \ref{fig:relu1-center-non-center}, \ref{fig:center-non-center-relu-sgd-transfer} and \ref{fig:center-non-center-relu-adam-transfer}. With this modification, we find empirically that the HP transfer prescription we first derived for linear networks works also for non-linear models. As we explain in App. \ref{SI-sec:centering} this mean-subtraction, which we refer as ``centering'', still gives a valid learned energy function. 
    In Sec. \ref{sec:empirics} we provide empirical results on non-linear DenseAMs.

    \item In the important setting of $\sigma = \text{softmax}$, we find analytically and empirically that SGD often suffers from numerical instability at large $N,K,P,B$. This stems from the fact that softmax differentially amplifies weight gradients from the two occurrences of model weights $W$. In contrast, Adam is scale-invariant and we show empirically that a simple prescription for scaling $s_1,s_2,\eta_W$ leads to HP transfer across $N,K, P,B$ in the proportional scaling regime \eqref{eq:proportional-regime}. See Figures \ref{fig:softmax-adam-center-non-center-transfer-and-mse}, \ref{fig:transfer-softmax-sgd-vs-adam} as well as Sec. \ref{sec:empirics} for more empirical results on softmax DenseAMs.
    Full details of the softmax analysis are in App. \ref{SI-sec:softmax}. 
    
    \item While our analytic computations assume isotropic data, we find empirically that our HP transfer prescriptions work surprisingly well for an-isotropic and realistic data. The core challenge here is to understand how to relate across different $N$ families of anisotropic data-generating processes (for isotropic data they are simply related by a scaled projection to a lower dimensional space). See Figures \ref{fig:transfer-anisotropic-relu} and \ref{fig:transfer-mnist}.
\end{itemize}

\noindent \textbf{Structure for Rest of Article.} The remainder of this article is organized as follows. Section \ref{sec:empirics} focuses on the practical aspects of this work. We detail and validate empirically our proposal for how to obtain reliable HP transfer across input/output dimension, width, and dataset size in DenseAMs. Section \ref{sec:DAM} provides a brief introduction to DenseAMs and explains how our scaling procedure is compatible with the learned energy. Section \ref{sec:theory} contains an overview of our technical analysis.

\section{HP Transfer Prescription for Shallow DenseAMs}\label{sec:empirics}

In this section we summarize (Sec. \ref{sec:param-summary}) and present empirical validation of HP transfer (Sec. \ref{sec:param-empirics}) for our proposed  parameterization (i.e. scalings of $s_1,s_2$ and learning rates with input/output dimension $N$, hidden layer width $K$, dataset size $P$, and batch size $B$). 

\subsection{Parameterization}\label{sec:param-summary} 

To achieve HP transfer, it is important to make the following structural modification to the DenseAM when using either SGD or Adam (compare with \eqref{eq:f-def})
\begin{equation}\label{eq:f-def-centered}
     f(x) =f_W(x)= s_2 W^T\sigma_C \big[s_1 W g(x) + b\big]+ c,
\end{equation}
where
\[
\sigma_C:= C\circ \sigma \circ C,\quad C = I_K - \frac{1}{K}{\bf 1}_K{\bf 1}_K^T.
\]
Notice that the matrix $C$ (which stands for ``centering'') is the projection in the latent space $\R^K$ to the orthogonal complement of the all ones vector ${\bf 1}$. Hence, $\sigma_C$ removes the layer-wise mean from both the pre and post activations. As we explain in App. \ref{SI-sec:centering}, neglecting to center $\sigma$ can lead to numerical instability and a lack of transfer due to model divergence (see Figures \ref{fig:relu1-center-non-center}, \ref{fig:center-non-center-relu-sgd-transfer} and \ref{fig:center-non-center-relu-adam-transfer}). Our parametrization is summarized in Table (\ref{tab:results}). 

\begin{table}[h]
    \centering
\begin{center}
\small
\renewcommand{\arraystretch}{1.2}
\begin{tabular}{|l|l|l|l|l|l|l|l|}
\hline
\textbf{Scaling regime} 
& \textbf{Act} 
& \textbf{Opt.} 
& \(\boldsymbol{s_1}\) 
& \(\boldsymbol{s_2}\)  
& \(\boldsymbol{\eta_W}\)
& \(\boldsymbol{\eta_b}\) 
& \(\boldsymbol{\eta_c}\) \\
\hline
\multirow{3}{*}{\parbox{3.8cm}{Proportional scaling \eqref{eq:proportional-regime} }} 
& ReLU\(^{p}\) & 
SGD  &
\(1/\sqrt{N}\) & 
\(1/\sqrt{K}\) & 
 \(\eta_0\,K\) &
 \(\eta_{b,0}\) &
 \(\eta_{c,0}\) \\
\cline{2-8}
& ReLU\(^{p}\) & 
Adam & 
\(1/\sqrt{N}\) & 
\(1/ \sqrt{K}\) & 
\(\eta_0\) &
 \(\eta_{b,0}\) &
 \(\eta_{c,0}\) \\
\cline{2-8}
& Softmax  
& \parbox{1.1cm}{Adam} 
& \(1/\sqrt{N}\) 
& \(\sqrt{K}\) 
& \parbox{1.1cm}{\(\eta_0\)} &
 \(\eta_{b,0}\) &
 \(\eta_{c,0}\) \\
\hline

\multirow{3}{*}{\parbox{3.8cm}{Scale $K$, Fix $N,P$ }} 
& ReLU\(^{p}\) & 
SGD  &
$1/\sqrt{N}$ & 
\(1/{K}\) & 
 \(\eta_0\,K\) &
 \(\eta_{b,0}\) &
 \(\eta_{c,0}\) \\
\cline{2-8}
& ReLU\(^{p}\) & 
Adam & 
 $1 / \sqrt{N}$ & 
\(1/ {K}\) & 
\(\eta_0\) &
 \(\eta_{b,0}\) &
 \(\eta_{c,0}\) \\
\cline{2-8}
& Softmax  
& \parbox{1.1cm}{Adam} 
& $1 / \sqrt{N}$
& 1
& \parbox{1.1cm}{\(\eta_0\)} &
 \(\eta_{b,0}\) &
 \(\eta_{c,0}\) \\
\hline
\end{tabular}
\end{center}

    \caption{Our proposed parameterization for scales $s_1,s_2$ and learning rates. In all experiments in which biases are trainable we set $\eta_{b,0}=\eta_{c,0}=\eta_0$ and sweep only the effective learning rate $\eta_0$.
    }
    \label{tab:results}
\end{table}

\begin{figure}
    \centering
    \includegraphics[width=0.87\linewidth]{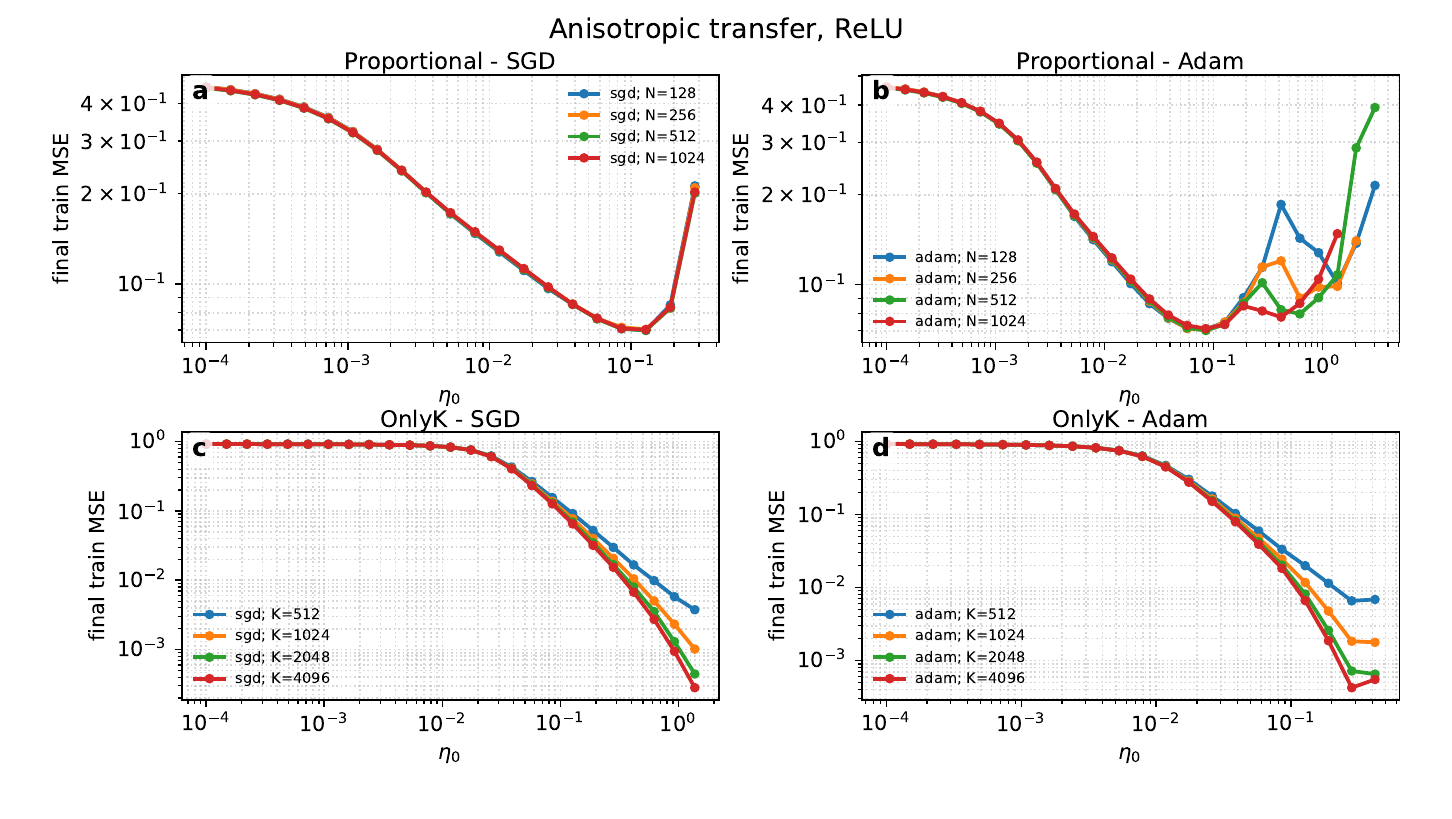}
    \caption{Learning rate transfer for SGD and Adam in the proportional regime \eqref{eq:proportional-regime} (top row), with $\kappa = 2, \rho=5, \beta=0.1$, and the width-only scaling (bottom row), with $N=128, P=256, \beta=0.1$ for ReLU DenseAMs. Models are trained to minimize denoising error \eqref{eq:loss} for anisotropic inputs $x_\alpha \sim \mathcal N(0,D)$, where $D$ is a diagonal matrix with $i$-th entry proportional to $i^{-2/5}$ and trace $N$.}
    \label{fig:transfer-anisotropic-relu}
\end{figure}

\noindent Note that our parameterization covers two distinct but related scaling regimes: 
\begin{itemize}
    \item \textbf{Simultaneous model and data scaling.} Here $N,K,P,B$ scale together in fixed proportion (see \eqref{eq:proportional-regime}). As we explain in Sec. \ref{sec:theory}, this leads not only to HP transfer but, at least empirically, to a collapse of training dynamics across scale. To enter this regime in practice one must have a method of changing the input dimension for the training data. In the simple case of denoising Gaussian covariates $y^\mu = x^\mu +\text{noise}$ with $x^\mu \sim \mathcal N(0,I_N)$ a simple projection suffices. For vision datasets, one can imagine up/down sampling images. See e.g. Figure \ref{fig:transfer-anisotropic-relu} for transfer across scales for up/down projections for anisotropic data.
    \item \textbf{Model only scaling.} Here we fix the data-related dimensions $N,P$ and vary model width $K$. In this regime, we do not expect a collapse of model dynamics, except that when $K\gg N,P$ we expect the dynamics will be similar to the setting where $\kappa= K/N\gives \infty$. The existence of this well-posed limiting dynamics is a key rationale for why we (and most prior work on $\mu$P-type HP transfer) expect HP transfer in this setting. See Figure \ref{fig:transfter-K-only-regime}.
\end{itemize}  

\subsection{Empirical Validation of HP Transfer}\label{sec:param-empirics}
In this section we report on a range of HP transfer experiments for shallow DenseAMs. Our goal is to demonstrate the utility of our parameterization and also some pitfalls and open questions. \\

\noindent \textit{Linear DenseAMs.} We begin with the setting of linear DenseAMs \eqref{eq:f-def} in which $\sigma = \mathrm{id}, g=\tanh$. Our main findings are as follows:
\begin{itemize}
    \item \textbf{Learning rate transfer.} We observe learning rate transfer in both the proportional regime and the $K$-only regime when training linear DenseAMs on isotropic data (see Figures \ref{fig:linear_SGD}, \ref{fig:transfter-K-only-regime}). 
    \item \textbf{Dynamical collapse.} For $N,K,P,B$ in the proportional regime \eqref{eq:proportional-regime} and isotropic data we find that not only learning rate transfer but that many internal statistics of training dynamics are constant across scale. We leave it as an interesting problem for future work to characterize these dynamics (a similar problem was solved without tied weights in \cite{bordelon2025deeplinearnetworktraining}). See Figure \ref{fig:linear_SGD}. 
\end{itemize}
In most $\mu$P-type approaches to HP transfer the parameterization that satisfies Desiderata 1-3 from Sec. \ref{sec:approach} in linear networks also gives rise to HP transfer in non-linear models. However, as we will see in the next section, this is not quite true for DenseAMs, where modifications to the architecture and optimizer are needed to adapt to non-linear DenseAMs.\\
\begin{figure}
    \centering
    \includegraphics[width=0.87\linewidth]{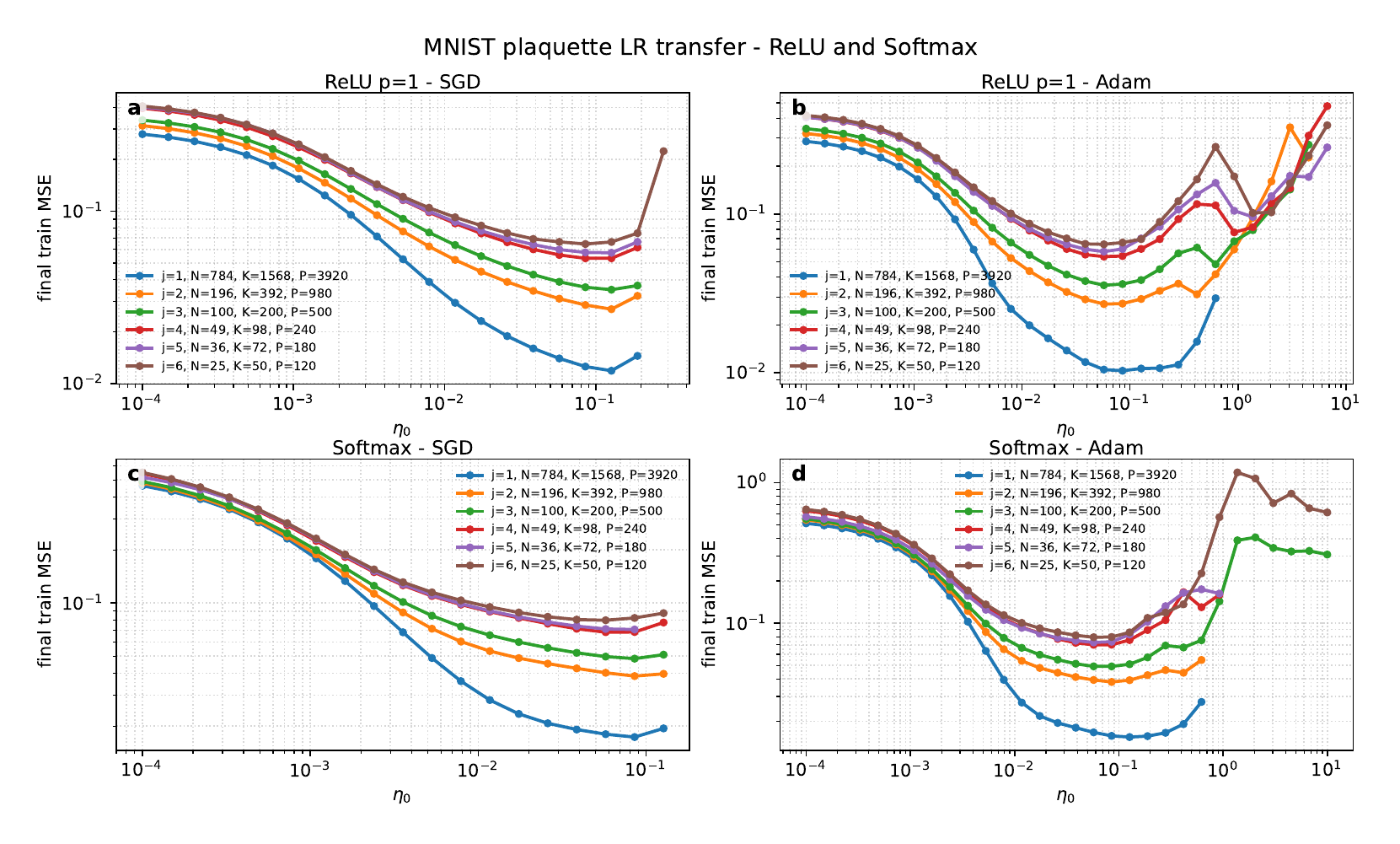}
    \caption{HP transfer for DenseAMs trained on MNIST using ReLU or softmax activations and SGD or Adam in the proportional regime \eqref{eq:proportional-regime} with $\kappa=2, \rho=5,\beta=0.1$ trained for 256 epochs. We vary input dimension by a  plaquette of size $j$ coarse-graining as described in Appendix \ref{sec:mnist}.}
    \label{fig:transfer-mnist}
\end{figure}

\noindent \textit{DenseAMs with ReLU-type activations.} The next model we consider is a DenseAM \eqref{eq:f-def} with 
\[
g= \text{tanh},\qquad \sigma \in \set{C_p\cdot \mathrm{ReLU}^p,\, p\geq 1},\qquad C_p^2 =\frac{1}{2}(2p-1)!!,
\]
where the normalizing constants are chosen so that $\mathbb E[\sigma^2(z)]=1$, and $g$ is chosen as a bounded function to keep the energy function \eqref{Eq: two-layer system energy} bounded. Such DenseAMs have favorable theoretical properties that allow them to memorize $\mathrm{poly}(K)$ training examples \cite{krotov2016dense}. Our main finding are as follows:
\begin{itemize}
    \item \textbf{Centering is necessary.} Even with $p=1$ non-linear ReLU DenseAMs give  HP transfer in both the proportional \eqref{eq:proportional-regime} and fixed $N,P$ regimes but only if one centers the pre and post-activations as in \eqref{eq:f-def-centered}. 
    See Figures \ref{fig:center-non-center-relu-sgd-transfer}, \ref{fig:center-non-center-relu-adam-transfer}, for the numerics and Sec. \ref{SI-sec:centering} for theoretical analysis of the effects of centering. 
    \item \textbf{Transfer for large $\mathrm{ReLU}$ powers.} At large values of $p$ Adam still exhibits LR transfer, while SGD training is often unstable at large learning rates. Adam remains stable even at $p=10$. See Figure \ref{fig:transfer-relu-p-large-p}. 
    \item \textbf{Dynamical collapse.} Empirically, for small values of $p$ (i.e. $p=1,2$) the internal statistics of centered non-linear DenseAMs give not only HP transfer but also consistent dynamics of internal network statistics. See Figures \ref{fig:dynamics-relu-sgd-dz}, \ref{fig:dynamics-collapse-relu-sgd-detailed}.
\end{itemize}

DenseAMs with ReLU-type activations considered here have a simple property: order $1$ pre-activations lead to order $1$ post-activations. This is no longer true for softmax non-linearities and necessitates a somewhat different treatment as we explain next.\\

\noindent \textit{Sofmax DenseAMs.} The final model we consider are DenseAMs \eqref{eq:f-def} with softmax non-linearity. As with DenseAMs with $\mathrm{ReLU}^p$ activations we center pre/post-activations  and find:
\begin{itemize}
    \item \textbf{SGD can fail to give transfer.} As shown in Figure \ref{fig:transfer-softmax-sgd-vs-adam} we find HP transfer is possible with Adam but not SGD. We offer a principled explanation for this phenomenon in App. \ref{SI-sec:softmax}.
    \item \textbf{Dynamical collapse.} Empirically, the internal statistics of softmax DenseAMs trained with Adam give not only HP transfer but also consistent dynamics of internal network statistics for centered softmax non-linearity. See Figures \ref{fig:softmax-adam-center-non-center-transfer-and-mse}, \ref{fig:softmax-dynamics-isntability}.
\end{itemize}

\section{Dynamics and energy of DenseAM}\label{sec:DAM}
We follow the general formulation and notation of DenseAMs with hidden units by Krotov and Hopfield \cite{krotovlarge}. The network is defined by two sets of neurons -- $N$ feature neurons $x_i$, and $K$ hidden neurons $h_\alpha$. 
The global energy function is given by 
\begin{equation} \label{Eq: two-layer system energy}
    E = \frac{1}{s_1}\sum\limits_{\alpha=1}^K (h_\alpha - b_\alpha)\sigma_\alpha - L_h + \frac{1}{s_2}\sum\limits_{i=1}^N (x_i - c_i)g_i - L_x ,
\end{equation}
where we assume that the Lagrangian for the feature neurons, $L_x$, is neuron-wise and the Lagrangian for the hidden neurons, $L_h$, can be neuron-wise or contrastive (resulting in a softmax activation function) 
\begin{equation}
\notag    L_x = \sum\limits_{i=1}^N G(x_i), \ \ \ \ \ L_h^\text{neuron-wise} = \sum\limits_{\alpha=1}^K \Phi(h_\alpha), \ \ \ \ \ L_h^\text{contrastive} = \frac{1}{\beta} \log\bigg[ \sum\limits_{\alpha=1}^K \exp(\beta h_\alpha)\bigg] .
\end{equation}
The activation functions are defined as derivatives of the Lagrangians 
\begin{equation}
\notag    \sigma_\alpha = s_1 \frac{\partial L_h}{\partial h_\alpha},  \ \ \ \ \  g_i = s_2 \frac{\partial L_x}{\partial x_i}.
\end{equation}
The biases $b_\alpha$ and $c_i$ are assumed to be constant, and the positive scalars $s_1$ and $s_2$ are introduced for the most flexible HP transfer. 

The total time-derivative of the energy function is given by 
\begin{align*}
        \frac{dE}{dt} =& \frac{1}{s_1}\sum\limits_{\alpha=1}^K\dot{h}_\alpha\sigma_\alpha + \frac{1}{s_1}\sum\limits_{\alpha,\gamma=1}^K (h_\alpha-b_\alpha) \frac{\partial \sigma_\alpha}{\partial h_\gamma} \dot{h}_\gamma - \sum\limits_{\alpha=1}^K \frac{\partial L_h}{\partial h_\alpha} \dot{h}_\alpha  \\
        &+ \frac{1}{s_2}\sum\limits_{i=1}^N\dot{x}_ig_i + \frac{1}{s_2}\sum\limits_{i=1}^N (x_i-c_i) \frac{\partial g_i}{\partial x_i} \dot{x}_i - \sum\limits_{i=1}^N \frac{\partial L_x}{\partial x_i} \dot{x}_i \\ & -\sum\limits_{\alpha,\gamma=1}^K\sum\limits_{i=1}^N \dot{h_\gamma} \frac{\partial \sigma_\alpha}{\partial h_\gamma} W_{\alpha i} g_i - \sum\limits_{\alpha=1}^K\sum\limits_{i=1}^N  \sigma_\alpha W_{\alpha i} \frac{\partial g_i}{\partial x_i}\dot{x}_i  \\ &=
        \frac{1}{s_1}\sum\limits_{\alpha,\gamma=1}^K \dot{h_\gamma }\frac{\partial \sigma_\alpha}{\partial h_\gamma} \Big[ h_\alpha-b_\alpha - s_1\sum\limits_{i=1}^N W_{\alpha i} g_i \Big] + \frac{1}{s_2}\sum\limits_{i=1}^N \dot{x_i}\frac{\partial g_i}{\partial x_i} \Big[ x_i-c_i - s_2\sum\limits_{\alpha=1}^K W_{\alpha i} \sigma_\alpha \Big] \\ 
        &= - \tau_h \sum\limits_{\alpha,\gamma=1}^K \dot{h_\gamma }\frac{\partial^2 L_h}{\partial h_\gamma\partial h_\alpha} \dot{h}_\alpha - \tau_x \sum\limits_{i=1}^N \dot{x_i}\frac{\partial^2 L_x}{\partial x_i^2} \dot{x}_i \leq 0 .
\end{align*}
The last equality uses the dynamical equations for the neurons 
\begin{equation*}
    \tau_h \dot{h}_\alpha = -h_\alpha + s_1 \sum\limits_{i=1}^N W_{\alpha i} g_i + b_\alpha,\qquad 
    \tau_x \dot{x}_i = -x_i + s_2 \sum\limits_{\alpha=1}^K W_{\alpha i} \sigma_\alpha + c_i,
\end{equation*}
and the inequality sign holds if the Lagrangians are convex (for neuron-wise Lagrangians this condition means that the activation function increases monotonically). 

Now consider the adiabatic limit, so that hidden neurons evolve on a much faster time scale compared to feature neurons. Mathematically, this means that $\tau_h\rightarrow 0$. In this limit, the first equation can be solved for the activities of the hidden neurons, assuming that they are in the steady-state for any value of feature neurons
\begin{equation} \label{Eq: adiabatic steady-state}
    h_\alpha = s_1 \sum\limits_{i=1}^N W_{\alpha i} g_i + b_\alpha.
\end{equation}
This allows us to write down an effective theory for feature neurons only. The dynamical equations in this limit give 
\begin{equation} \label{Eq: effective dynamics continuous time}
    \tau_x \dot{x}_i = -x_i + s_2 \sum\limits_{\alpha=1}^K W_{\alpha i} \sigma_\alpha\Big[ s_1 \sum\limits_{j=1}^N W_{\gamma j} g_j + b_\gamma\Big] + c_i,
\end{equation}
and the effective energy -- substitute Eq. (\ref{Eq: adiabatic steady-state}) into the definition of energy (\ref{Eq: two-layer system energy}) -- is given by 
\begin{equation}
    E^\text{eff} = - L_h \Big[ s_1 \sum\limits_{j=1}^N W_{\gamma j} g_j + b_\gamma\Big] + \frac{1}{s_2} \sum\limits_{i=1}^N g_i (x_i-c_i) - L_x .
\end{equation}
Finally, in order to simulate the continuous dynamical system on a GPU we need to discretize Eq. (\ref{Eq: effective dynamics continuous time}). For simplicity, we assume that the time discretization step is the same as the time constant of the dynamics $dt=\tau_x=1$. This leads to 
\begin{equation}
    x^{t+1}_i - x^t_i = -x_i^t + s_2 \sum\limits_{\alpha=1}^K W_{\alpha i} \sigma_\alpha\Big[ s_1 \sum\limits_{j=1}^N W_{\gamma j} g_j^t + b_\gamma\Big] + c_i,
\end{equation}
which coincides with the update Eq. (\ref{eq:f-def}) written in matrix notations. To make it easier to connect with the literature on HP transfer, we use the notation $f(x)$ to denote the updated vector $x$. Thus, Eq. (\ref{eq:f-def}) defines a one-step update of the network's state towards the direction of energy minimization. It has all aspects of weight-sharing as general DenseAM networks, and also incorporates general biases for feature and hidden neurons.

\section{Deriving HP Transfer in DenseAMs}\label{sec:theory}
In this section we present an overview of our approach to HP transfer in DenseAMs starting with some notation in Sec. \ref{sec:notation}.

\subsection{Notations}\label{sec:notation}
Recall that the dense associative memory networks we study have the form 
\begin{equation}
\label{eq:f-def-2}
    f(x) =f_W(x)= s_2 W^T\sigma\big[s_1 W g(x) + b\big]+ c,\quad x,g(x),c\in \R^N, \, b \in \R^K, \, W\in \R^{K\times N}, 
\end{equation}
and that our goal is to optimize $W, b, c$ through minimizing an empirical MSE 
\begin{equation}
\label{eq:loss-2}
\mathcal L(W):=\frac{1}{2B}\sum_{\mu=1}^B \norm{f_W(x^\mu + \epsilon^\mu) - x^\mu}^2,\qquad \epsilon^\mu \sim N(0, \sigma_\epsilon^2 I_{N\times N})
\end{equation}
using either mini-batch SGD or Adam with batch size $B$ and learning rates $\eta_W, \eta_b, \eta_c$ for the weights and biases, respectively. To efficiently analyze DenseAM training dynamics we introduce several abbreviations. First, let us denote for every data point  $x^\mu$
\[
g^{\mu} = g(x^{\mu}), \quad z^{\mu} = s_1 W g^{\mu}, \quad  f^{\mu} = f(x^{\mu}), \quad r^{\mu} = f^{\mu} - y^{\mu} .
\]
Arranging all such vectors across a training batch as columns of a matrix, gives the following matrices 
\begin{equation}
\label{eq:matrix-notation}
 G \in \mathbb{R}^{N \times B}, \quad Z \in \mathbb{R}^{K \times B},\quad S = \sigma(Z) \in \mathbb{R}^{K \times B}, \quad F \in \mathbb{R}^{N \times B}, \quad R = F - Y \in \mathbb{R}^{N \times B}.
\end{equation}

\subsection{Approach to HP Transfer}\label{sec:approach}
Our approach to HP transfer modifies in the style of \cite{Dey25Dont} the $\mu$P heuristic from \cite{yang2020feature}. Namely, we seek scalings of $s_1, s_2, \eta_W, \eta_b, \eta_c$ as functions of $N,K,P,B$ to satisfy: 
\begin{itemize}
    \item \textbf{Desideratum 1.} (Stability) We require that, at initialization, the entries of $Z, F$ be order $1$ in the sense that their mean remains $O(1)$ and their variance is $\Theta(1)$ as $N,K\gives \infty$. 
    \item \textbf{Desideratum 2.} (Maximality) We require that for any fixed $T\geq 1$ the change $\Delta Z, \Delta F$ in each component of $Z,F$ after each of the first $T$ steps of optimization is order $1$ as $N,K,P,B\gives \infty$. 
    \item \textbf{Desideratum 3.} (Balance) At first order in the learning rate $\Delta Z,\Delta F$ decompose as
    \begin{equation}\label{eq:balanced}
        \Delta Z = \Delta_b Z +\sum_{\ell=1}^2 \Delta_W^{(\ell)} Z ,\qquad \Delta F = \Delta_{b} F + \Delta_{c}F + \sum_{\ell,\ell'=1,2}\Delta_W^{(\ell,\ell')}F 
    \end{equation}
    corresponding to the change in $Z,F$ coming from changing the biases $b,c$ and from which appearance of the tied weights $W$ is being updated in the linearization of $Z,F$ with respect to $W$. We require that each component of each term in this decomposition is order $1$ as $N,K,P,B\gives \infty$. See \eqref{eq:DW-decomposition}-\eqref{eq:dF-in-terms-dW}.
\end{itemize}
Since the scale $s_1,s_2$ are variable,\footnote{A priori one should allow for a \textit{additional} scales $s_b$, $s_c$ to for the biases $b,c$. A direct computation very similar to the one below for choosing $s_1,s_2$ suggests we should set $s_b=s_c=1$.} we can and shall assume a standard Gaussian initialization: 
    \begin{align}
        b_i, c_j, W_{ij}\sim \mathcal N(0,1)\quad \text{iid}\qquad \text{at initialization.}
    \end{align}
Relative to prior work on HP transfer, analyzing DenseAMs involves several subtleties: 
\begin{itemize}
    \item HP transfer techniques typically do not address  growing output dimension $N$. 
    \item Due to the presence of tied weights, the output and input dimensions agree. So by considering the setting of growing output dimension we are forced to consider growing input dimension and hence families of data-generating processes indexed by $N$. For our core derivations we will assume that training inputs $x^\mu$ follow an isotropic Gaussian law $\mathcal N(0,I_N)$.
    \item HP transfer techniques are relatively unexplored for models with tied weights. The requirement for a \textit{balanced} parameterization as in \eqref{eq:balanced} --particularly the requirement that $\Delta_W^{_{(1)}}F$ and $\Delta_W^{_{(2)}}F$ match in scale -- is therefore novel. 
\end{itemize}
We leave a complete analytic description of training dynamics, even in the case of linear activations, to future work. Instead, we aim to perform a minimal set of careful computations that allow us to guess how $s_1, s_2, \eta_W, \eta_b, \eta_c$ should scale. To do so, we record here several exact formulas, such as the expressions for $\Delta Z, \Delta F$, starting with the weight and bias gradient of the MSE loss \eqref{eq:loss}:
\begin{align}
    \label{eq:loss-grad}
    \frac{\partial \mcL}{\partial W} &= \frac{s_2}{B} S R^{\top} + \frac{s_1 s_2 }{B} \left( S' \odot (W R) \right) G^{\top},\quad 
\frac{\partial \mathcal L}{\partial b}
= \frac{s_2}{B} \left( S' \odot (W R) \right) \mathbf{1},\quad 
\frac{\partial \mathcal L}{\partial c}
= \frac{1}{B} R \mathbf{1}.
\end{align}
The two terms in $\partial_W \mcL$ in \eqref{eq:loss-grad} come from the change of the outer layer (first term) and the change of the hidden features (second term), which are summed since first and second layer weights are tied.  For SGD\footnote{Adam is analyzed in App. \ref{SI-sec:adam}}, the changes in the biases, weights, pre-activations, and outputs are
\begin{align}
\label{eq:DW-decomposition}
\Delta W &= -\eta\frac{s_2}{B}  S R^{\top} - \eta \frac{s_1 s_2}{B} \left( S' \odot (W R) \right) G^{\top}   := \Delta^{(1)} W + \Delta^{(2)} W\\
\label{eq:Dc}\Delta b &= -\eta_b \frac{s_2}{B} \left( S' \odot (W R) \right) \mathbf{1}, \quad \Delta c = -\eta_c \frac{1}{B} R\mathbf{1}\\
\label{eq:Z-decomposition}
\Delta_W Z &= s_1 \Delta W G,\quad \Delta_b Z = \Delta b {\bf 1}^\top \\
\label{eq:dF-in-terms-dW}
\Delta_W F &= s_2 \Delta W^{\top} S + s_1 s_2 W^{\top} \left( S' \odot ( \Delta W G ) \right),\quad \Delta_c F = \Delta c \mathbf{1}^\top.
\end{align}
As alluded to in \eqref{eq:balanced} $\Delta Z$  contains two terms coming from the tied weights and $\Delta F$ has four such terms after substituting \eqref{eq:DW-decomposition} (see Eqs. \eqref{SI-eq:dZ-expanded} and \eqref{SI-eq:dF-4-terms}). For linear activations $\sigma(z)=z$ we calculate the expected magnitudes $Z, F, \Delta Z, \Delta F$ in App. \ref{SI-sec:linear-sgd}. The result is that Desiderata 1 - 3 are met only when $N,K,P,B$ are proportional as in \eqref{eq:proportional-regime} and we set 
\begin{equation}\label{eq:sgd-linear-hp}
     s_1 = \frac{s_{1,0}}{\sqrt{N}}, \quad s_2 = \frac{s_{2,0}}{\sqrt{K}}, \quad \eta_W= K\eta_{0},\quad \eta_b = \eta_{b,0},\quad \eta_c=\eta_{c,0}
\end{equation}
where $s_{1,0}, s_{2,0}, \eta_{0}, \eta_{b,0},\eta_{c,0}$ are order $1$ constants. Our derivations suggest that with the scaling \eqref{eq:sgd-linear-hp} we should expect not only transfer of $\eta_0$ across $N,K,P,B$ but also that the training dynamics of the learned representation $Z,$ output $F$, and loss $\mathcal L$ are consistent across scale. 
See Figures \ref{fig:dynamics-relu-sgd-dz}, \ref{fig:dynamics-collapse-relu-sgd-detailed}. In the Appendix we explain how to extend the parameterization \eqref{eq:sgd-linear-hp} to include the Adam optimizer (App. \ref{SI-sec:adam}), and to non-linear activations (App. \ref{SI-sec:centering}) including softmax (App. \ref{SI-sec:softmax}).

\begin{figure}
    \centering
    \includegraphics[width=1\linewidth]{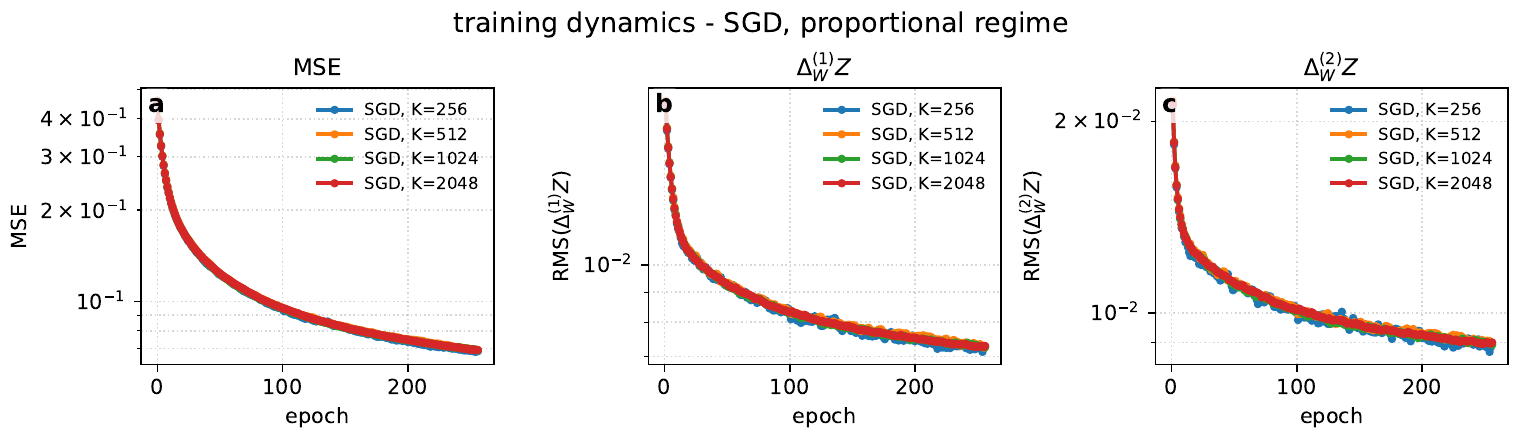}
    \caption{Dynamical consistency across scales for MSE (left), $\Delta_W^{(1)}Z, \, \Delta_W^{(2)}Z$ for a ReLU DenseAM trained with SGD in the proportional regime \eqref{eq:proportional-regime} with $\kappa=5, \rho=2, \beta=0.1$, and $\eta_0 = 0.005$.}
    \label{fig:dynamics-relu-sgd-dz}
\end{figure}

\section*{Conclusion} 
In this article we have initiated the study of HP transfer for DenseAMs. We proposed and validated empirically a novel parameterization for the simple DenseAMs \eqref{eq:f-def}, which are generalizations of Hopfield networks. We have identified and addressed several practically important considerations, such as the need to center pre and post-activations and the numerical instability SGD with softmax activations. While our analysis was based on precise computations, we did not develop a full description of training dynamics at infinite scale for DenseAMs. We leave this, as well as extensions to more modern architectures, to future work.

\bibliography{bibliograhy}
\bibliographystyle{unsrt}

\newpage
\appendix

\section{Model and definitions}
\label{SI-sec:model}

We study dense associative memory (DenseAM) networks of the form
\begin{equation}
\label{eq:SI-uncentered-model}
    f(x) =f_W(x)= s_2 W^\top\sigma\big[s_1 W g(x) + b\big]+ c,\quad x,g(x),c\in \R^N, \, b \in \R^K, \, W\in \R^{K\times N}. 
\end{equation}
Our goal is to train a denoiser using the empirical MSE
\begin{equation}
\label{eq:SI-loss}
\mathcal L(W):=\frac{1}{2B}\sum_{\mu=1}^B \norm{f_W(x^\mu + \epsilon^\mu) - x^\mu}^2,\qquad \epsilon^\mu \sim N(0, \sigma_\epsilon^2 I_{N\times N}),
\end{equation}
by either mini-batch SGD or Adam with batch size \( B \) and learning rates \( \eta_W, \eta_b, \eta_c \).

Let us introduce the following notations.
For every data point  $x^\mu$ we denote
\begin{equation}
\label{eq:single-sample-notation}    
g^{\mu} = g(x^{\mu}), \quad z^{\mu} = s_1 W g^{\mu}, \quad  f^{\mu} = f(x^{\mu}), \quad r^{\mu} = f^{\mu} - y^{\mu} .
\end{equation}
For a mini-batch \(\mathcal B\) of size \(B\), we collect the corresponding vectors as columns:
\begin{equation}
\label{eq:SI-matrix-notation}
G \in \mathbb{R}^{N \times B}, \quad Z \in \mathbb{R}^{K \times B},\quad S = \sigma(Z) \in \mathbb{R}^{K \times B}, \quad F \in \mathbb{R}^{N \times B}, \quad R = F - Y \in \mathbb{R}^{N \times B}.
\end{equation}
Then, the mini-batch loss is
\begin{equation}
\label{SI-eq:batch-loss}
    \mcL_{\mathcal B} = \frac{1}{2B}\norm{F-Y}_F^2 = \frac{1}{2B}\norm{R}_F^2.
\end{equation}

We define the proportional scaling regime by the following
\begin{equation}
\label{SI-eq:proportional-regime}
    K=\kappa N, \qquad P=\rho N, \qquad B=\beta P=\rho_B N, \qquad \rho_B\equiv \beta\rho.
\end{equation}

\subsection{Uncentered and Centered DenseAMs}

We refer to the model in Eq.~\eqref{eq:SI-uncentered-model} as the uncentered model. 
Let us next introduce the centered model which is crucial for achieving HP transfer in case of DenseAMs with activations $\sigma$ that have nonzero mean, such as $\mathrm{ReLU}^p$ and softmax. 
We define the centering operator
\begin{equation}
\label{SI-eq:C-def}
    C=I_K-\frac{1}{K}\mathbf{1}_K\mathbf{1}_K^\top, \qquad\tilde W = C W, \qquad \tilde b = C b,
\end{equation}
and we denote $\tilde W, \tilde b$ the centered weights.
The centered forward pass is given by
\begin{equation}
\label{eq:centered-model}
f(x)  = s_2 \tilde{W}^{\top} \sigma\left(s_1 \tilde{W} g(x) + \tilde b \right) + c =
s_2 W^\top C \, \sigma\left( s_1 C W g(x) + C b \right) + c .
\end{equation}
Thus the preactivations are centered across hidden units, while the post-activation contribution to the output is projected by \(C\).
This convention that replaces $W, b$ by $\tilde W = CW , \tilde b = C b$ in \eqref{eq:SI-uncentered-model} preserves the energy formulation discussed in Sec. \ref{sec:DAM}.

Some of the formulas below are written for the uncentered elementwise model.
For the centered model, the same formulas apply with \(W, b\) replaced by \(\tilde W=CW, \tilde b = Cb\), together with an additional left projection \(C\) when converting gradients with respect to \(\tilde W, \tilde b\) into gradients with respect to the underlying parameters \(W, b\).
This is made explicit in App. \ref{SI-sec:centered-projected-gd}.

\section{One-step updates for elementwise activations}
\label{SI-sec:one-step-updates}

In this section \(\sigma\) is applied elementwise on $z \in \R^K$.  
Following \eqref{eq:SI-matrix-notation}, it is useful to denote the matrix
\[
    S' = \sigma'(Z)\in\R^{K\times B}.
\]
For now we use the uncentered model \eqref{eq:SI-uncentered-model}, and the centered modification \eqref{eq:centered-model} is described in App. \ref{SI-sec:centered-projected-gd}.

\subsection{Weight-gradient decomposition}
\label{SI-sec:weight-gradient}

The derivative of the \(i\)-th component of the output \(f_i^\mu\) in \eqref{eq:SI-uncentered-model} with respect to \(W_{ab}\) is
\begin{align*}
\label{SI-eq:df-dW-entry}
    \frac{\partial f_i^\mu}{\partial W_{ab}} &= s_2\frac{\partial}{\partial W_{ab}} \sum_{j=1}^K W_{ji}\sigma(z_j^\mu)
    = s_2\delta_{ib}\sigma(z_a^\mu) + s_1s_2 W_{ai}\sigma'(z_a^\mu)g_b^\mu.
\end{align*}
Therefore, summing over a mini-batch yields
\begin{equation*}
    \left[\nabla_W \mcL_{\mathcal B}\right]_{ab} = \frac{1}{B}\sum_{\mu=1}^B \sum_{i=1}^N r_i^\mu \frac{\partial f_i^\mu}{\partial W_{ab}} = \frac{s_2}{B}\sum_{\mu=1}^B  r_b^\mu S_{a\mu} + \frac{s_1s_2}{B} \sum_{\mu=1}^B  S'_{a\mu} \left(\sum_{i=1}^N W_{ai}r_i^\mu\right) G_{b\mu}.
\end{equation*}
In matrix notation,
\begin{equation}
\label{SI-eq:grad-W}
    \nabla_W\mcL_{\mathcal B} = \frac{s_2}{B}S R^\top + \frac{s_1s_2}{B} \left(S'\odot (W R)\right)G^\top.
\end{equation}
Thus an SGD step with weight learning rate \(\eta_W\) has
\begin{equation}
\label{SI-eq:dW}
    \Delta W = -\eta_W\nabla_W\mcL_{\mathcal B}= \Delta^{(1)} W+\Delta^{(2)} W,
\end{equation}
where
\begin{equation}
  \label{SI-eq:dW1-dW2}
  \Delta^{(1)} W = -\eta_W\frac{s_2}{B}SR^\top, \qquad \Delta^{(2)} W= -\eta_W\frac{s_1s_2}{B} \left(S'\odot (WR)\right)G^\top.
\end{equation}
The two terms correspond to the two appearances of the tied weight matrix \(W\): \(\Delta^{(1)}W\) comes from the outer occurrence \(W^\top S\), while \(\Delta^{(2)}W\) comes from the hidden preactivation \(WG\).

\subsection{Hidden representation update}
\label{SI-sec:hidden-update}

The preactivations are given by
\[
    Z=s_1WG+b\mathbf{1}_B^\top,
\]
and therefore, their \( W \) update is
\begin{equation}
\label{SI-eq:dZ-exact}
    \Delta_W Z=s_1\Delta W G.
\end{equation}
Substituting \(\Delta W=\Delta^{(1)}W+\Delta^{(2)}W\) from \eqref{SI-eq:dW1-dW2},
\begin{equation}
  \label{SI-eq:dZ-decomp}
  \Delta_W Z = \Delta_W^{(1)} Z+\Delta_W^{(2)} Z, \qquad \Delta_W^{(\ell)} Z=s_1\Delta^{(\ell)}W G.
\end{equation}
Explicitly, 
\begin{equation}
\label{SI-eq:dZ-expanded}
    \Delta_W^{(1)} Z = -\eta_W\frac{s_1 s_2}{B}S R^\top G , \qquad 
    \Delta_W^{(2)} Z = - \eta_W\frac{s_1^2 s_2}{B} \left(S'\odot (WR)\right)(G^\top G).
\end{equation}

\subsection{Output update}
\label{sec:output-update}

To first order in \(\Delta W\), the output update is given by
\begin{equation}
\label{SI-eq:dF-in-terms-dW}
    \Delta_W F = s_2(\Delta W)^\top S + s_1s_2W^\top\left(S'\odot(\Delta W G)\right) + O\left(\norm{\Delta W}^2\right).
\end{equation}
Substituting \(\Delta W=\Delta^{(1)}W+\Delta^{(2)}W\) from \eqref{SI-eq:dW1-dW2} gives four terms:
\begin{align}
  \label{SI-eq:dF-decomp}
  \Delta_W F &= s_2(\Delta^{(1)}W)^\top S + s_2(\Delta^{(2)}W)^\top S \\
             & + s_1 s_2W^\top\left(S'\odot(\Delta^{(1)}WG)\right) + s_1 s_2 W^\top\left(S'\odot(\Delta^{(2)}WG)\right) + O\left(\norm{\Delta W}^2\right) \nonumber\\
             &:= \Delta_W^{(1,1)} F + \Delta_W^{(1,2)} F + \Delta_W^{(2,1)} F + \Delta_W^{(2,2)} F + O\left(\norm{\Delta W}^2\right).
  \nonumber
\end{align}
Explicitly,
\begin{subequations}
\label{SI-eq:dF-4-terms}
\begin{align}
\label{SI-eq:dF-11} \Delta_W^{(1,1)} F &= -\eta_W\frac{s_2^2}{B}R S^\top S, \\
\label{SI-eq:dF-12} \Delta_W^{(1,2)} F &= -\eta_W\frac{s_1s_2^2}{B} G\left(\left(S'\odot(WR)\right)^\top S\right), \\
\label{SI-eq:dF-21} \Delta_W^{(2,1)} F &= -\eta_W\frac{s_1s_2^2}{B} W^\top\left(S'\odot\left(S(R^\top G)\right)\right), \\
\label{SI-eq:dF-22} \Delta_W^{(2,2)} F &= -\eta_W\frac{s_1^2s_2^2}{B} W^\top\left[ S'\odot\left(\left(S'\odot(WR)\right)(G^\top G)\right) \right].
\end{align}
\end{subequations}
The term \(\Delta_W^{(1,1)} F\) is the direct output change from \(\Delta^{(1)}W\), \(\Delta_W^{(1,2)} F\) is the direct output change from \(\Delta^{(2)}W\), \(\Delta_W^{(2,1)} F\) is the feature-induced output change from \(\Delta^{(1)}W\), and \(\Delta_W^{(2,2)} F\) is the feature-induced output change from \(\Delta^{(2)}W\).

\subsection{Bias gradients}

The gradient with respect to the inner bias \(b\) is given by
\begin{equation}
\label{SI-eq:grad-b}
    \nabla_b\mcL_{\mathcal B} = \frac{s_2}{B} \left(S'\odot(WR)\right)\mathbf{1}_B.
\end{equation}
Therefore, the inner bias update is and its effect on the preactivations are
\begin{equation}
\label{SI-eq:db}
    \Delta b = -\eta_b\frac{s_2}{B} \left(S'\odot(WR)\right)\mathbf{1}_B, \qquad \Delta_b Z=\Delta b\,\mathbf{1}_B^\top.
\end{equation}
For the output bias \(c\),
\begin{equation}
\label{SI-eq:grad-c}
    \nabla_c\mcL_{\mathcal B} = \frac{1}{B}R\mathbf{1}_B,
\end{equation}
and therefore its update and effect on the output updates are
\begin{equation}
\label{SI-eq:dc}
    \Delta c = -\eta_c\frac{1}{B}R\mathbf{1}_B, \qquad \Delta_cF=\Delta c\,\mathbf{1}_B^\top.
\end{equation}

\subsection{Centered model as projected gradient descent}
\label{SI-sec:centered-projected-gd}

For the centered model \eqref{eq:centered-model}, the loss depends on the trainable matrix \(W\) only through
\[
    \tilde W=CW.
\]
Therefore, by the chain rule,
\begin{equation}
\label{SI-eq:centered-grad}
    \nabla_W\mcL_{\mathcal B} = C\nabla_{\tilde W}\mcL_{\mathcal B}.
\end{equation}
An SGD update of \(W\) gives
\[
    \Delta W=-\eta_W C\nabla_{\tilde W}\mcL_{\mathcal B},
\]
and hence (using the fact that $C^2 = C$, recall \eqref{SI-eq:C-def})
\begin{equation}
\label{SI-eq:centered-dWtilde}
    \Delta\tilde W = C\Delta W = -\eta_W C^2\nabla_{\tilde W}\mcL_{\mathcal B} = -\eta_W C\nabla_{\tilde W}\mcL_{\mathcal B}.
\end{equation}
Thus training the centered model is a projected gradient descent on the column-centered weight subspace \(\mathbf{1}_K^\top \tilde W=0\).
Namely, the MSE loss \eqref{eq:SI-loss} of the centered network \eqref{eq:centered-model} is invariant under the transformation $W \mapsto W + a {\bf 1}_K^\top$ for any vector $a \in \R^N$.

In practice, the uncentered formulas above carry over after replacing \(W\) by \(\tilde W\) and projecting with \(C\).  For example, the term \(\Delta_W^{(1,1)} F\) in Eq. \eqref{SI-eq:dF-11} becomes
\begin{equation}
\label{SI-eq:centered-dF-11}
    \Delta_W^{(1,1)} F = -\eta_W\frac{s_2^2}{B} R S^\top C S.
\end{equation}
The replacement
\[
    S^\top S \mapsto S^\top C S
\]
is the key change responsible for removing the mean-induced spike discussed in Section~\ref{SI-sec:centering}.

\section{Initialization scaling of hidden preactivations and outputs}
\label{SI-sec:initialization}

Throughout this section we work at initialization.  
The entries of \(W\) are iid with mean zero and variance one, and the entries of \(G\) are taken to have mean zero and variance \(v_g=O(1)\).  
For the centered model, \(W\) should be replaced by \(\tilde W=CW\).  
This introduces weak correlations between hidden rows and changes \(K\) to \(K-1\), but it does not change any scaling exponent.

\subsection{Hidden preactivation scale}
\label{SI-sec:s1-scaling}

Ignoring the bias, the preactivation matrix is
\[
    Z= s_1 W G.
\]
Each entry is a sum of \(N\) independent terms,
\[
    Z_{k\mu} = s_1 \sum_{i=1}^N W_{ki} G_{i\mu}.
\]
Thus, the variance of an entry is
\[
    \Var(Z_{k\mu})=s_1^2 N v_g.
\]
To keep \(Z_{k\mu}=O(1)\), we choose
\begin{equation}
\label{SI-eq:s1-scaling}
    s_1\sim \frac{1}{\sqrt N}.
\end{equation}

\subsection{Output scale for linear activations}
\label{SI-sec:s2-scaling}

We next analyze the output scaling for the simple case of linear activations $\sigma(z)=z$.
Ignoring biases, the output matrix is given by
\begin{equation*}
\label{SI-eq:F-linear}
    F=s_2W^\top Z=s_1s_2W^\top W G.
\end{equation*}
A Frobenius-norm computation gives the typical entry size.  For a collection of \(B\) samples,
\begin{equation*}
\label{SI-eq:F-frob-start}
    \frac{\E{\norm{F}_F^2}}{NB} = \frac{s_1^2s_2^2}{NB} \E{\tr\left(G^\top (W^\top W)^2G\right)} = \frac{s_1^2s_2^2v_g}{N} \E{\tr\left((W^\top W)^2\right)}.
\end{equation*}
For \(W\in\R^{K\times N}\) with iid standard Gaussian entries a direct computation yields,  
\begin{equation*}
\label{SI-eq:WTW-second}
    \E{\tr\left((W^\top W)^2\right)} = KN(K+N+1).
\end{equation*}
Therefore, with \(s_1\sim N^{-1/2}\) from Eq. \eqref{SI-eq:s1-scaling},
\begin{equation}
\label{SI-eq:F-frob-final}
    \frac{\E{\norm{F}_F^2}}{NB} = s_1^2 s_2^2 v_g K (K+N+1) = s_2^2 v_g K \left(1 + \frac{K}{N} + \frac{1}{N} \right).
\end{equation}
Thus, in the proportional regime \(K=\kappa N\), to keep the output entries order one, we ought to set
\begin{equation}
\label{SI-eq:s2-proportional-linear}
    s_2\sim \frac{1}{\sqrt K}, \qquad \text{proportional regime}.
\end{equation}
In contrast, if \(K\to\infty\) while \(N, B, P\) are fixed (see App. \ref{sec:K-only-scaling}), then the leading behavior of \eqref{SI-eq:F-frob-final} is
\[
    \frac{\E{\norm{F}_F^2}}{NB} \sim s_2^2 v_g \frac{K^2}{ N},
\]
and hence, to maintain it order one, we set
\begin{equation}
\label{SI-eq:s2-onlyK-linear}
    s_2\sim \frac{1}{K},\qquad \text{only-$K$ regime}.
\end{equation}

\section{Linear SGD scaling in the proportional regime}
\label{SI-sec:linear-sgd}

We specialize to the case of linear activation,
\[
    \sigma(z)=z, \qquad S=Z, \qquad S'=\mathbf{1}_{K\times B}.
\]
The purpose of this section is to determine the scaling of \(\eta_W\) in the proportional regime \eqref{SI-eq:proportional-regime}.
We focus on the network-dependent part of the residual by replacing \(R\) with \(F\) in Eq. \eqref{SI-eq:dW1-dW2}.
The omitted target-dependent terms have the same or smaller scaling under the isotropic assumptions and do not change the scaling exponents.

\subsection{Hidden update}
\label{SI-sec:dZ}

Using \(S=Z=s_1WG\), Eqs. \eqref{SI-eq:dZ-decomp}, \eqref{SI-eq:dZ-expanded} become
\begin{equation}
\label{SI-eq:dZ-linear}
    \Delta_W Z = -\eta_W\frac{s_1^2s_2}{B} W\left(GR^\top+RG^\top\right)G.
\end{equation}
Replacing \(R\) by \(F=s_1s_2W^\top WG\), we obtain two terms:
\begin{align}
\label{SI-eq:dW1Z-F-expanded}
    \Delta_W^{(1)} Z_F &= -\eta_W\frac{s_1^3s_2^2}{B} WGG^\top W^\top WG, \\
\label{SI-eq:dW2Z-F-expanded}
    \Delta_W^{(2)} Z_F &= -\eta_W\frac{s_1^3s_2^2}{B} WW^\top WGG^\top G.
\end{align}

The first term can be written conveniently in terms of \(Z=s_1WG\),
\begin{equation}
\label{SI-eq:dW1Z-F-Z-form}
    \Delta_W^{(1)} Z_F = -\eta_W\frac{s_2^2}{B} Z Z^\top Z.
\end{equation}
Hence, its Frobenius norm is given by
\begin{equation}
\label{SI-eq:dW1Z-F-frob}
    \norm{\Delta_W^{(1)} Z_F}_F^2 = \left(\eta_W\frac{s_2^2}{B}\right)^2 \tr\left((Z^\top Z)^3\right).
\end{equation}
Conditioned on \(G\), the hidden rows of \(Z\) are independent Gaussian vectors with covariance
\[
    \Sigma_G=s_1^2G^\top G\in\R^{B\times B}.
\]
Indeed, the \( k \)-th row of \( Z \) is given by
\[
z_k^{\top} = s_1 w_k^{\top} G
\]
and since the rows of \( W \) are independent, the rows of \( Z \) are also independent.
As \( W_{ij} \sim \mathcal{N}(0, 1) \), the row \( z_k^{\top} | G \sim \mathcal{N}(0, \Sigma_G) \), where \( \Sigma_G = s_1^2 G^{\top} G \).
(Note that \( z_k^{\top} \) is approximately gaussian when \( N \) is large, even if \( W_{ij} \) are not gaussian, by the central limit theorem).
Therefore
\[
Z^{\top} Z = \sum_{k=1}^{K} z_k z_k^{\top} \in \R^{B \times B}
\]
is a Wishart matrix with \(K\) degrees of freedom and covariance \(\Sigma_G\).  
Going back to Eq. \eqref{SI-eq:dW1Z-F-frob}, we need the third Wishart moment
\begin{equation*}
\label{SI-eq:wishart-3}
    \E{\tr((Z^{\top} Z)^3)\mid G} = Kt_1^3 + 3K(K+1)t_1t_2 + (K^3+3K^2+4K)t_3, \qquad     t_j=\tr(\Sigma_G^j).
\end{equation*}
With \(s_1^2 = 1/N\), the empirical covariance \(\Sigma_G = G^\top G / N\) has Marchenko-Pastur moments in the proportional regime \eqref{eq:proportional-regime}, where $\rho_B = B / N$ is fixed
\begin{equation}
\label{SI-eq:MP-moments}
    t_j \sim Bv_g^j m_j(\rho_B), \qquad
    m_1=1, \qquad
    m_2=1+\rho_B, \qquad
    m_3=1+3\rho_B+\rho_B^2.
\end{equation}
Substitution of Eqs. \eqref{SI-eq:wishart-3}, \eqref{SI-eq:MP-moments} into \eqref{SI-eq:dW1Z-F-frob}, and using $\kappa = K / N$, gives 
\begin{equation}
\label{SI-eq:dW1Z-F-final}
    \frac{\E{\norm{\Delta_W^{(1)} Z_F}_F^2}}{KB} \sim (\eta_Ws_2^2)^2v_g^3 \left[ 1 + 3\kappa\left(1+\frac{1}{\rho_B}\right) + \kappa^2\left(1+\frac{3}{\rho_B}+\frac{1}{\rho_B^2}\right) \right].
\end{equation}

For the second term $\Delta_W^{(2)} Z_F$, it is convenient to define
\begin{equation}
\label{SI-eq:AW-AG}
    A_W=\frac{1}{K}W^\top W \in \R^{N \times N},\qquad
    A_G=\frac{1}{B}GG^\top \in \R^{N \times N}.
\end{equation}
Then, Eq. \eqref{SI-eq:dW2Z-F-expanded} becomes
\begin{equation}
\label{SI-eq:dW2Z-F-frob}
    \norm{\Delta_W^{(2)} Z_F}_F^2 = \left(\eta_W\frac{s_1^3s_2^2}{B}\right)^2 B^3K^3 \tr(A_W^3A_G^3).
\end{equation}
Since \(A_W\) is orthogonally invariant and independent of \(A_G\), rotational averaging gives
\[
    \E{\tr(A_W^3A_G^3)} = \frac{1}{N}\E{\tr(A_W^3)}\E{\tr(A_G^3)}.
\]
Using the corresponding Marchenko-Pastur moments for fixed $\kappa = K / N, \rho_B = \beta P / N = B / N$ while $K, N, B, P \to \infty$,
\begin{align*}
    \E{\tr(A_W^3)} &\sim N\left(1+\frac{3}{\kappa}+\frac{1}{\kappa^2}\right), \\
    \E{\tr(A_G^3)} &\sim Nv_g^3 \left(1+\frac{3}{\rho_B}+\frac{1}{\rho_B^2}\right),
\end{align*}
we find Eq. \eqref{SI-eq:dW2Z-F-frob} scales as
\begin{equation}
\label{SI-eq:dW2Z-F-final}
    \frac{\E{\norm{\Delta_W^{(2)} Z_F}_F^2}}{KB} \sim (\eta_Ws_2^2)^2v_g^3 \left(1+3\kappa+\kappa^2\right) \left(1+\frac{3}{\rho_B}+\frac{1}{\rho_B^2}\right).
\end{equation}
Both hidden-update terms $\Delta_W^{(1)} Z_F, \Delta_W^{(2)} Z_F$, in Eqs. \eqref{SI-eq:dW1Z-F-final}, \eqref{SI-eq:dW2Z-F-final}, have the same powers of $\kappa$ and $\rho_B$ and only differ in their coefficients. 
Therefore they have the common scaling
\begin{equation}
\label{SI-eq:dZ-scaling-summary}
    \frac{\E{\norm{\Delta_W^{(\ell)} Z}_F^2}}{KB} \sim (\eta_Ws_2^2)^2h_\ell(\kappa,\rho_B), \qquad \ell =1,2
\end{equation}
where \(h_\ell ( \kappa, \rho_B)\) is a function of order one at fixed $\kappa, \rho_B$.

\subsection{Output update}
\label{SI-sec:dF}

We next calculate the magnitudes of the output updates.

Substituting \(S=Z\), \(S'=\mathbf{1}_{K\times B}\), and replacing \(R\) with \( F \) in Eqs.~\eqref{SI-eq:dF-11}--\eqref{SI-eq:dF-22}, and using \eqref{SI-eq:AW-AG}, yields
\begin{subequations}
\label{SI-eq:dF-id-activation}
\begin{align}
\label{SI-eq:dF-11-F}
    \Delta_W^{(1,1)} F_F &= -\eta_Ws_2^3s_1^3K^2 A_WA_GA_WG, \\
\label{SI-eq:dF-12-F}
    \Delta_W^{(1,2)} F_F &= -\eta_Ws_2^3s_1^3K^2 A_GA_W^2G, \\
\label{SI-eq:dF-21-F}
    \Delta_W^{(2,1)} F_F &= -\eta_Ws_2^3s_1^3K^2 A_WA_GA_WG = \Delta_W^{(1,1)} F_F, \\
\label{SI-eq:dF-22-F}
    \Delta_W^{(2,2)} F_F &= -\eta_Ws_2^3s_1^3K^2 A_W^2A_GG.
\end{align}
\end{subequations}
Consider the last term's Frobenius norm
\begin{align}
\label{SI-eq:dFD-frob}
    \norm{\Delta_W^{(2,2)} F_F}_F^2 &= \left(\eta_Ws_2^3s_1^3K^2\right)^2 B\tr(A_W^4A_G^3),
\end{align}
which is very similar to the norm of \(\Delta_W^{(2)} Z_F\) in Eq. \eqref{SI-eq:dW2Z-F-frob}, and so we use the same calculation steps.
Rotational averaging gives
\begin{align}
\label{SI-eq:dFD-final}
    \E{\norm{\Delta_W^{(2,2)} F_F}_F^2} &\sim \left(\eta_Ws_2^3s_1^3K^2\right)^2 BNv_g^3 m_4\left(\frac{1}{\kappa}\right) m_3\left(\frac{1}{\rho_B}\right),
\end{align}
where, $m_3$ is given in \eqref{SI-eq:MP-moments}, and the fourth Marchenko-Pastur moment is
\[
    m_4(a)=1+6a+6a^2+a^3.
\]
Normalizing, and using $s_1 = 1/\sqrt{N}$, Eq. \eqref{SI-eq:dFD-final} becomes (compare with Eq. \eqref{SI-eq:dW2Z-F-final})
\begin{equation}
    \frac{\E{\norm{\Delta_W^{(2,2)} F_F}_F^2}}{NB} \sim \left(\eta_W s_2^2\right)^2 (s_2^2 K) v_g^3 m_4 \left({\kappa}\right) m_3\left(\frac{1}{\rho_B}\right).
\end{equation}

The other terms of $\Delta_W F$ in Eqs. \eqref{SI-eq:dF-id-activation} involve alternating products of \(A_W\) and \(A_G\).
Explicit calculation of the other terms yields same powers of $\kappa$ and $\rho_B$, but with different coefficients.
This is the same behavior we have found above for $\Delta_W^{(1)} Z$ and $\Delta_W^{(2)} Z$. Indeed, compare \eqref{SI-eq:dW1Z-F-final} with \eqref{SI-eq:dW2Z-F-final} and see they have the same powers of $\kappa$ and $\rho_B$.
Hence all four output-update terms in \eqref{SI-eq:dF-id-activation} have the common scaling
\begin{equation}
\label{SI-eq:dF-scaling-summary}
    \frac{\E{\norm{\Delta_W^{(\ell,\ell')} F}_F^2}}{NB} \sim \left(\eta_W s_2^2\right)^2 (s_2^2 K) h_{\ell, \ell'}(\kappa,\rho_B).
\end{equation}
where \(h_{\ell, \ell'}(\kappa,\rho_B)\) is a function of order one at fixed $\kappa, \rho_B$.

\subsection{Resulting SGD prescription}

Recall our desiderata 1-3 in Sec. \ref{sec:approach}, where we require that (1) at initialization the entries of $Z, F$ are order one, (2) the updates $\Delta Z, \Delta F$ are of order one, and (3) all terms of the updates, i.e., $\Delta_W^{(\ell)} Z, \Delta_W^{(\ell, \ell')} F$, are balanced.

Desideratum 1, analyzed in Secs. \ref{SI-sec:s1-scaling} and \ref{SI-sec:s2-scaling}, has yielded (see Eqs. \eqref{SI-eq:s1-scaling} and \eqref{SI-eq:s2-proportional-linear}) for the proportional regime,
\begin{equation}
\label{SI-eq:s1-s2-scaling-for-SGD-prescription}
    s_1\sim N^{-1/2},\qquad s_2\sim K^{-1/2}.
\end{equation}

Sections \ref{SI-sec:dZ} and \ref{SI-sec:dF}, dealing with Desiderata 2 and 3, have provided two scaling relations we need to satisfy. One in \eqref{SI-eq:dZ-scaling-summary} for $\Delta_W^{(\ell)} Z$ and the other in \eqref{SI-eq:dF-scaling-summary} for $\Delta_W^{(\ell, \ell')} F$.

Plugging \eqref{SI-eq:s1-s2-scaling-for-SGD-prescription} into \eqref{SI-eq:dZ-scaling-summary} we get 
\[
    \frac{\E{\norm{\Delta_W^{(\ell)} Z}_F^2}}{KB} \sim \left( \frac{\eta_W}{K} \right)^2.
\]
Thus \(\Delta_W^{(\ell)} Z=O(1)\) requires
\[
    \eta_W\sim K.
\]
Plugging \eqref{SI-eq:s1-s2-scaling-for-SGD-prescription} into \eqref{SI-eq:dF-scaling-summary},
\[
   \frac{\E{\norm{\Delta_W^{(\ell,\ell')} F}_F^2}}{NB} \sim \frac{\eta_W^2}{K^2}.
\]
Thus \(\Delta_W^{(\ell,\ell')} F=O(1)\) also requires
\[
    \eta_W\sim K.
\]
Therefore, for linear activations trained with SGD in the proportional regime, satisfying our desiderata requires
\begin{equation}
\label{SI-eq:linear-sgd-prescription}
    s_1\sim \frac{1}{\sqrt N}, \qquad s_2\sim \frac{1}{\sqrt K}, \qquad \eta_W\sim K.
\end{equation}
We find empirically that this prescription not only provides HP transfer, but also keeps dynamical collapse during training, see Fig. \ref{fig:linear_SGD}. 

\begin{figure}
    \centering
    \includegraphics[width=1\linewidth]{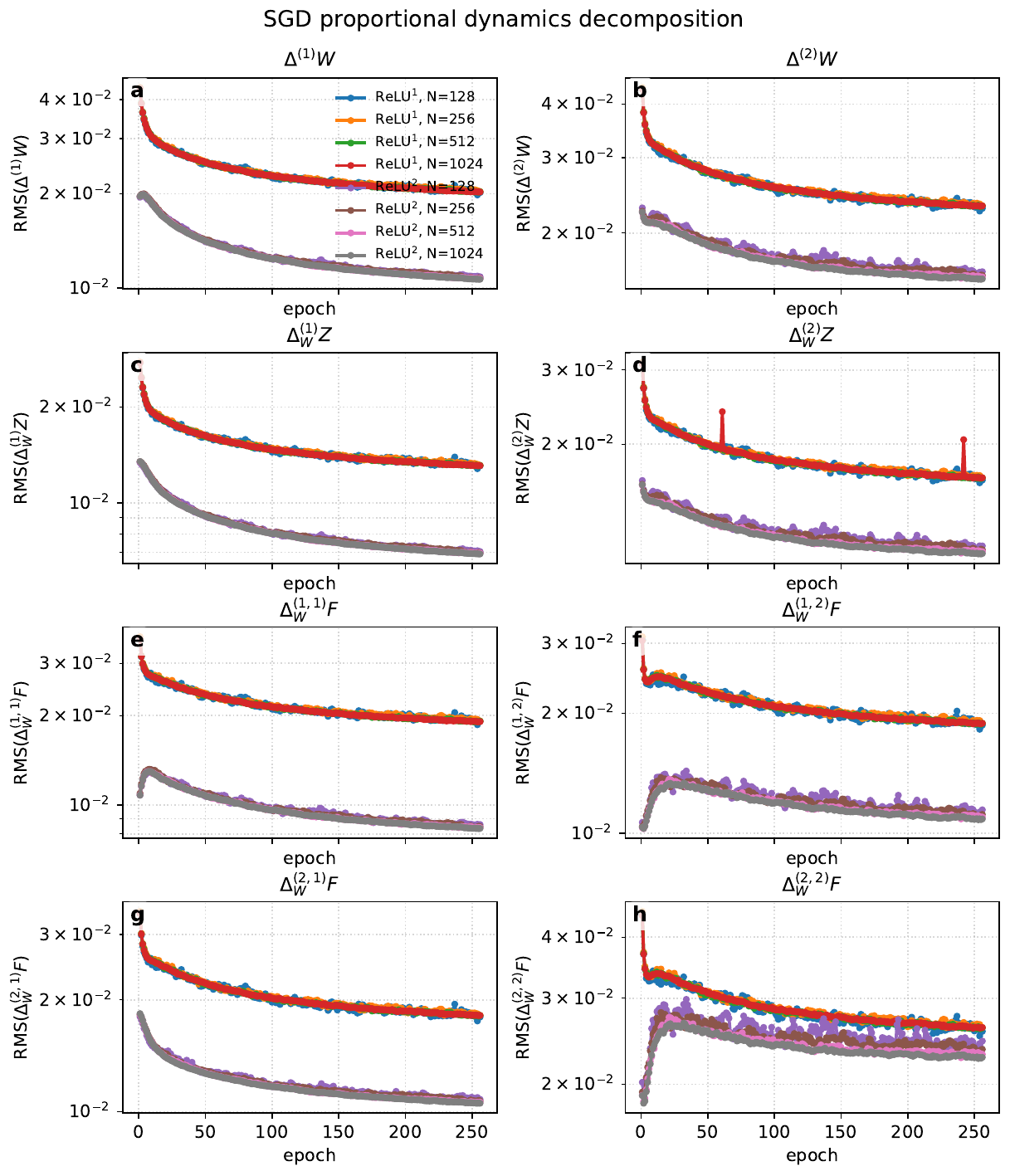}
    \caption{
    Training dynamics collapse for ReLU$^{p}$ DenseAMs trained with SGD in the proportional regime \eqref{eq:proportional-regime} with $\kappa=2, \rho=5, \beta=0.1$.
    We observe that the weight updates $\Delta^{(1)} W, \Delta^{(2)} W$ in Eq. \eqref{SI-eq:dW1-dW2}, the preactivations updates $\Delta_W^{(1)} Z, \, \Delta_W^{(2)} Z$ in Eq. \eqref{SI-eq:dZ-expanded}, and the output updates $\Delta_W^{(1,1)} F, \, \Delta_W^{(1,2)} F, \, \Delta_W^{(2,1)} F, \, \Delta_W^{(2,2)} F$ in Eqs. \eqref{SI-eq:dF-4-terms} all behave the same as $N$ increases.
    }
    \label{fig:dynamics-collapse-relu-sgd-detailed}
\end{figure}

\section{Nonlinear elementwise activations and centering}
\label{SI-sec:centering}

For nonlinear elementwise activations, the main additional issue is that nonzero-mean activations (such as $\mathrm{ReLU}^p$) create an outlying eigenvalue in the hidden-feature Gram matrix $S^\top S$.
This affects significantly the output update through the term (Eq.  \eqref{SI-eq:dF-11})
\[
    \Delta_W^{(1,1)} F = -\eta_W\frac{s_2^2}{B}R S^\top S.
\]

\subsection{Mean-induced spike}
\label{SI-sec:spike-nonzero-activation}

For each sample \(\mu\), decompose the hidden activation vector into its hidden-unit mean and a centered fluctuation
\begin{equation}
\label{SI-eq:sigma-bar-def}
    \bar\sigma_\mu=\frac{1}{K}\sum_{k=1}^K S_{k\mu},\qquad \tilde S_{k\mu}=S_{k\mu}-\bar\sigma_\mu.
\end{equation}
In matrix form,
\begin{equation}
\label{SI-eq:S-mean-decomp}
    S=\tilde S+\mathbf{1}_K\bar\sigma^\top,\qquad \tilde S^\top\mathbf{1}_K=0,
\end{equation}
and therefore the Gram matrix is decomposed as
\begin{equation}
\label{SI-eq:S-gram-decomp}
    S^\top S = \tilde S^\top\tilde S + K\bar\sigma\bar\sigma^\top.
\end{equation}
If the activation has nonzero mean \(m=\E{\sigma(z)}\neq 0\), then \(\bar\sigma\approx m\mathbf{1}_B\) (Eq. \eqref{SI-eq:sigma-bar-def}), so
\begin{equation}
\label{SI-eq:activation-spike}
   S^\top S \approx \tilde S^\top\tilde S +  Km^2\mathbf{1}_B\mathbf{1}_B^\top.
\end{equation}
This rank-one term has eigenvalue
\begin{equation}
\label{SI-eq:lambda-spike-elementwise}
    \lambda_{\rm spike}(S^\top S) \approx KBm^2.
\end{equation}
By contrast, if the centered fluctuations have order-one variance and weak correlations, the bulk eigenvalues of \(\tilde S^\top\tilde S\) are \(  \lambda_{\rm bulk} \left(S^\top S \right) = \lambda_{\rm bulk} \left(\tilde S^\top\tilde S \right) \sim O(K)\).  
The spike \eqref{SI-eq:lambda-spike-elementwise}  therefore scales differently from the bulk and prevents a single learning rate scaling from matching both contributions.
Indeed, in the proportional regime $B \sim K$, and $\lambda_{\rm spike} \sim K^2$ whereas $\lambda_{\rm bulk} \sim K$.
Setting $\eta_W$ to scale $\lambda_{\rm bulk}$ proportionally will cause $\lambda_{\rm spike}$ to grow too rapidly, and will cause an instability at large $N \sim K \sim P \sim B$.
See Fig. \ref{fig:center-non-center-relu-sgd-eigenvalues}(a,b).
On the other hand, setting $\eta_W$ to scale $\lambda_{\rm spike}$ proportionally will cause $\lambda_{\rm bulk}$ to grow too slowly and will slow down the training dynamics as the DenseAM scales up. 

To check the spike and bulk behavior, we train a DenseAM with ReLU activation using SGD. We monitor the largest eigenvalue of $S^\top S/K$, denoted by $\lambda_{\max} (S^\top S / K)$ which corresponds to the spike of $S$, and the second largest eigenvalue, denoted by $\lambda_2 (S^\top S / K)$, which corresponds to the bulk of $S$.
Indeed, we find that using prescription \eqref{SI-eq:linear-sgd-prescription}, the bulk shows consistent dynamics as we scale up the DenseAM (see Fig. \ref{fig:center-non-center-relu-sgd-eigenvalues}(b)), whereas the spike increases with scale $N$ (see Fig. \ref{fig:center-non-center-relu-sgd-eigenvalues}(a)).
That is why, as we show next, centering is necessary.

\subsection{Removing the spike by centering}

Recall the centering operator \eqref{SI-eq:C-def}
\[
    C=I_K-\frac{1}{K}\mathbf{1}_K\mathbf{1}_K^\top,
\]
which removes the problematic nonzero mean $\bar \sigma_\mu$ in \eqref{SI-eq:sigma-bar-def} exactly (using $C \mathbf{1}_K^\top = 0$)
\[
    CS = C \left( \tilde S+\mathbf{1}_K\bar\sigma^\top\right)= C \tilde S =\tilde S .
\]
Therefore, using $C^2 = C, C^\top = C$,
\[
    S^\top C S = \tilde S^\top\tilde S,
\]
and the rank-one spike \(K\bar\sigma\bar\sigma^\top\) is removed from the training dynamics.

\subsection{Centering the weights maintains the energy function of DenseAM}
\label{sec:centering}

In order to maintain the energy function of the DenseAM (see Sec. \ref{sec:DAM}), to incorporate the centering introduced above, we use the centered network in \eqref{eq:f-def-centered}.
Essentially, this means that we replace $W$ of the uncentered network \eqref{eq:f-def} by $\tilde W = CW$ (and we use $C^\top = C$).

\begin{figure}
    \centering
    \includegraphics[width=1\linewidth]{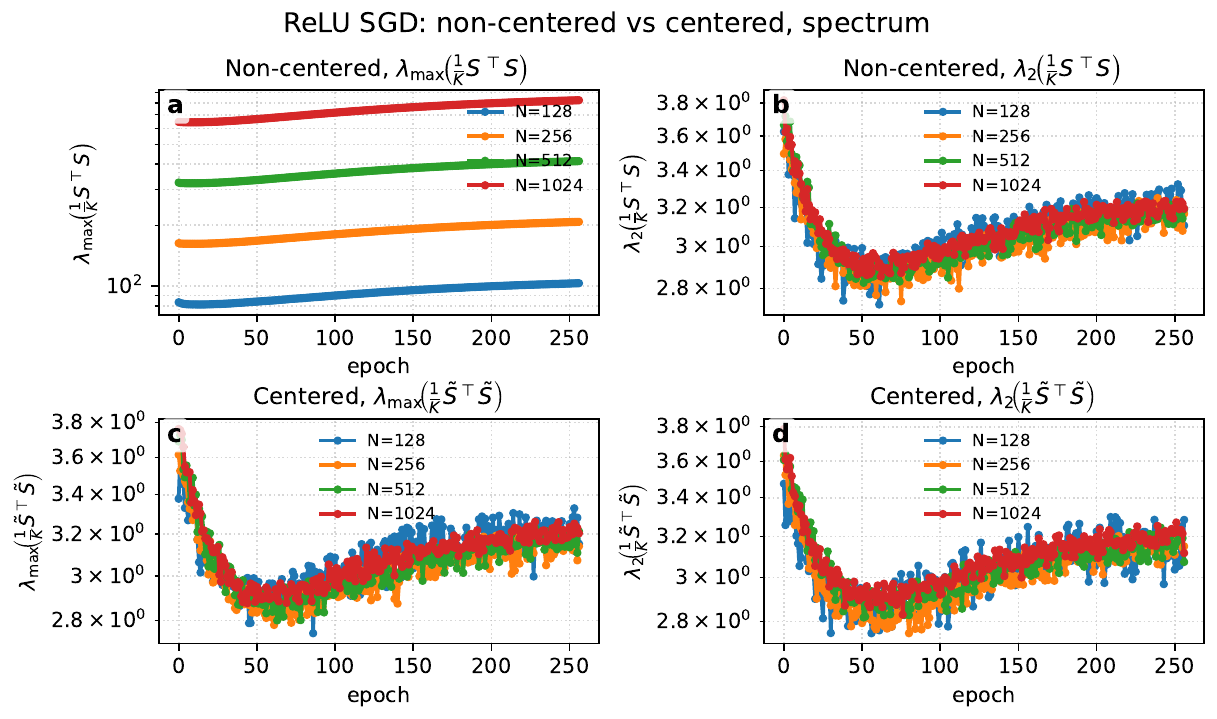}
    \caption{Spike and bulk behaviors for centered and non-centered ReLU DenseAM trained mini-batch SGD in the proportional regime \eqref{eq:proportional-regime} with scale factors $\kappa=2, \rho=5, \beta=0.1$ and $256$ epochs. 
    The left panels show the maximal eigenvalues $\lambda_{\max} (S^\top S / K)$ and $\lambda_{\max} (\tilde S^\top \tilde S / K)$, corresponding to the non-centered network and the centered network, respectively. These reflect the spike discussed in App. \ref{SI-sec:spike-nonzero-activation}. 
    The right panels show the second largest eigenvalues $\lambda_{2} (S^\top S / K)$ and $\lambda_{2} (\tilde S^\top \tilde S / K)$, corresponding to the non-centered network and the centered network, respectively.
    These reflect the bulk eigenvalues.
    In the non-centered network, the spike (panel (a)), and the bulk (panel (b)) do not have the same order of magnitude, and they scale differently. 
    In particular, the bulk behavior collapses as we scale up the DenseAM, whereas the spike keeps increasing, ruining the balance (desideratum 3 in Sec. \ref{sec:approach}).
    Centering fixes this issue.
    This can be seen in panels (c,d) which show the spike and the bulk for the centered DenseAM.
    The eigenvalues collapse well across scale, indicating the success of HP transfer depicted in Fig. \ref{fig:center-non-center-relu-sgd-transfer}.
    }
    \label{fig:center-non-center-relu-sgd-eigenvalues}
\end{figure}

\begin{figure}
    \centering
    \includegraphics[width=1\linewidth]{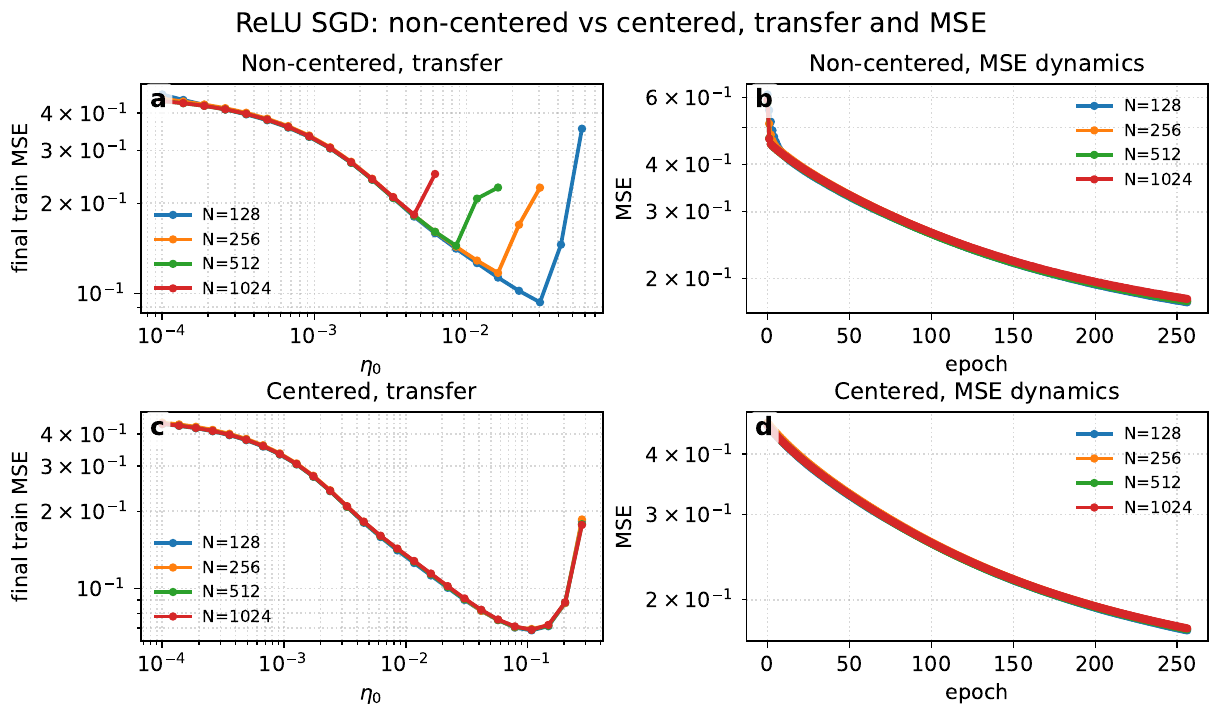}
    \caption{HP transfer for centered and non-centered ReLU DenseAM trained mini-batch SGD in the proportional regime \eqref{eq:proportional-regime} with scale factors $\kappa=2, \rho=5, \beta=0.1$ and $256$ epochs. 
    The left panels show the MSE loss after 256 training epochs as function of $\eta_0$ following the prescription in Table \ref{tab:results}, setting $s_1 =1/\sqrt{N}, s_2 =1/\sqrt{K}, \eta_W = \eta_0 K$. 
    Panel (a) shows that HP transfer is unstable for the non-centered DenseAM. 
    Indeed, increasing $N$ causes the DenseAM to become unstable at lower values of $\eta_0$, ruining transfer. 
    This happens due to the different scales of the spike and the bulk of the Gram matrix $S^\top S$, as shown in Fig. \ref{fig:center-non-center-relu-sgd-eigenvalues}.
    Centering, shown in panel (c), removes this mismatch in scales (see Fig. \ref{fig:center-non-center-relu-sgd-eigenvalues}(c,d)), and HP transfer is achieved.
    The right panels show the evolution of the MSE loss during training for $\eta_0 = 0.005$.
    The centered DenseAM (panel (d)) exhibits much better collapse across scales than the non-centered DenseAM (panel (b)).
    }
    \label{fig:center-non-center-relu-sgd-transfer}
\end{figure}

Since the centered network \eqref{eq:centered-model} sees $\tilde S$ instead of $S$, the spike is removed. 
In Figs. \ref{fig:center-non-center-relu-sgd-eigenvalues}, \ref{fig:center-non-center-relu-sgd-transfer}  we compare between centered and non-centered ReLU DenseAM. 
Figure \ref{fig:center-non-center-relu-sgd-eigenvalues}(a,c) shows the top eigenvalue (spike) of the appropriate Gram matrix, $S^\top S/K$ for the non-centered network, and $\tilde S^\top \tilde S / K$ for the centered network.
Panels (b,d) show the second largest eigenvalue (bulk) of these Gram matrices.
We observe that centering indeed makes the spike and the bulk of the same order, and moreover, induces dynamical collapse as the DenseAM is scaled in the proportional regime. 
Figure \ref{fig:center-non-center-relu-sgd-transfer}(b,d) shows the evolution of the MSE loss, and it can be seen that the centered DenseAM exhibits better dynamical collapse with respect to the non-centered DenseAM.

Figure \ref{fig:center-non-center-relu-sgd-transfer} also reveals the instability issue induced by the spike for a non-centered DenseAM.
In the non-centered case, the spike (shown in Fig. \ref{fig:center-non-center-relu-sgd-eigenvalues}(a)) increases as we scale up the DenseAM. 
This means that for any $\eta_0$ (which enters the learning rate as $\eta_W = \eta_0 K$, see Table \ref{tab:results}) increasing $K$, will cause the learning rate to be too large, and destabilize training. 
Panel (a) in Fig.  \ref{fig:center-non-center-relu-sgd-transfer} shows this: as $N\sim K$ increases, the final MSE loss increase rapidly for smaller $\eta_0$. 
Panel (c), which shows the final MSE loss for centered DenseAM, does not exhibit this issue.

\begin{figure}
    \centering
    \includegraphics[width=1\linewidth]{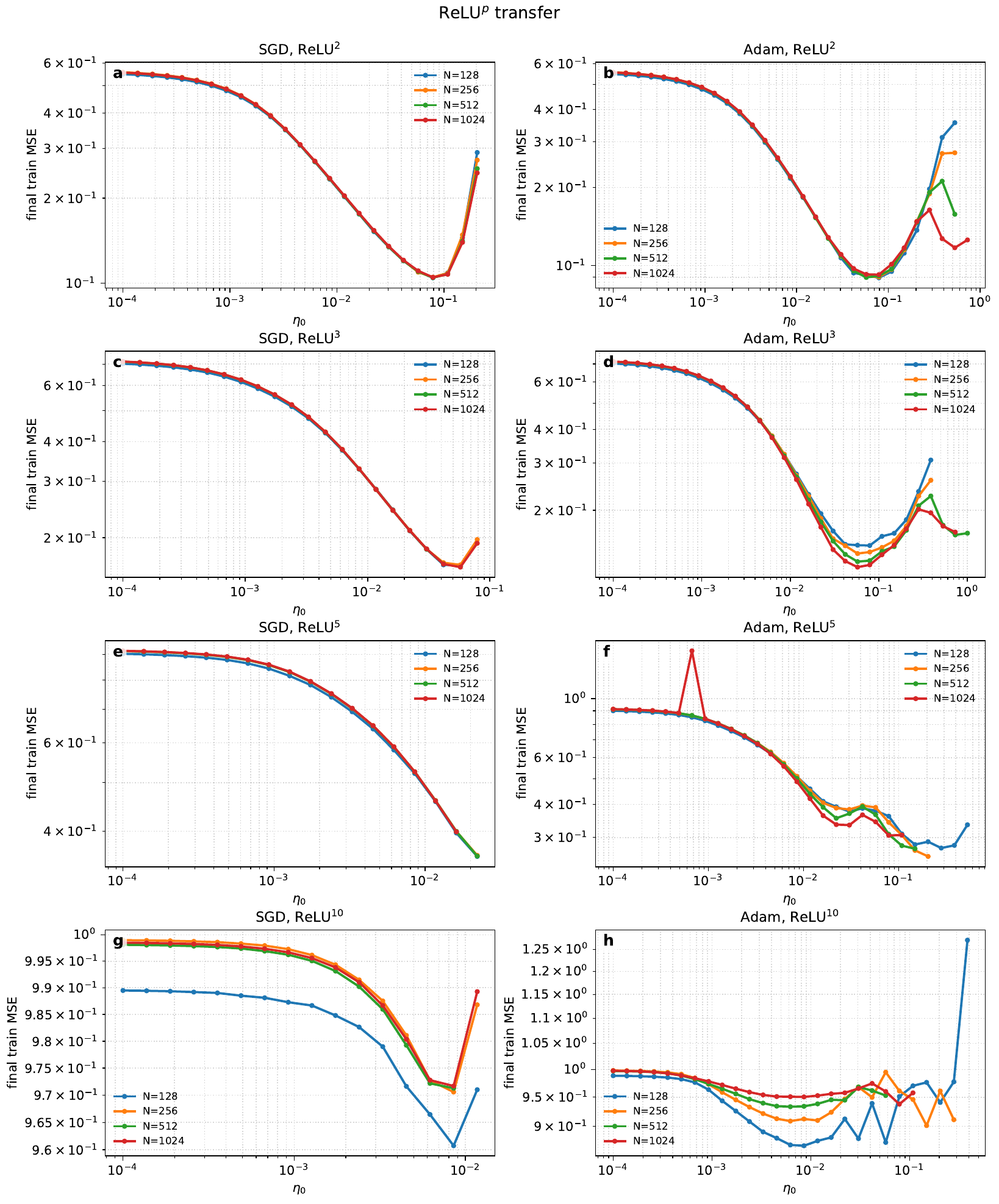}
    \caption{Learning rate transfer for a DenseAM with centered $\mathrm{ReLU}^p$ non-linearity trained by minimizing the denoising objective \eqref{eq:loss} using mini-batch SGD or Adam in the proportional regime \eqref{eq:proportional-regime} with scale factors $\kappa=2, \rho=5, \beta=0.1$ and $256$ epochs. As the nonlinearity $p$ increases, SGD often exhibits instability at larger learning rates, while Adam remains stable. 
    } 
    \label{fig:transfer-relu-p-large-p}
\end{figure}

Finally, in Fig. \ref{fig:transfer-relu-p-large-p} we show HP transfer for centered ReLU$^p$ DenseAMs with increasing powers of $p$. We observe that large powers destabilize HP transfer, but Adam handles larger $p$ better.

\section{Adam optimizer}
\label{SI-sec:adam}

\subsection{Mean-induced spike effects in Adam}
\label{SI-sec:adam-spike}

The spike discussed in Sec.~\ref{SI-sec:spike-nonzero-activation} appears directly in SGD through the term \(\Delta_W^{(1,1)}F\), because the SGD update is proportional to the gradient.
Adam changes this mechanism because it rescales the gradient entrywise.
Therefore the maximal eigenvalue of \(S^\top S\) does not enter Adam in exactly the same way.
However, the same nonzero mean of the post-activations in Eq.~\eqref{SI-eq:S-mean-decomp} can still produce a contribution to the Adam update and cause a mismatch in the scaling during training.

We use the gradient decomposition from Eq. \eqref{SI-eq:grad-W},
\begin{equation}
    \label{SI-eq:adam-grad-split}
    \nabla_W\mcL_{\mathcal B}=\nabla_W^{(1)}\mcL_{\mathcal B}+\nabla_W^{(2)}\mcL_{\mathcal B},
\end{equation}
with
\begin{equation}
    \label{SI-eq:adam-grad-terms}
    \nabla_W^{(1)}\mcL_{\mathcal B}=\frac{s_2}{B}S R^\top,\qquad \nabla_W^{(2)}\mcL_{\mathcal B}=\frac{s_1s_2}{B}\left(S'\odot(WR)\right)G^\top .
\end{equation}
We focus on \( \nabla_W^{(1)}\mcL_{\mathcal B} \) because it is the one that corresponds to \(\Delta_W^{(1,1)}F\) for SGD, which is responsible for the spike via the Gram matrix $S^\top S$.

Adam forms entrywise moving averages
\begin{equation}
    \label{SI-eq:adam-mv}
    M_t=\beta_1M_{t-1}+(1-\beta_1)\nabla_W\mcL_{\mathcal B,t},\qquad V_t=\beta_2V_{t-1}+(1-\beta_2)(\nabla_W\mcL_{\mathcal B,t})^{\odot 2},
\end{equation}
with \(M_0=V_0=0\). The bias-corrected moments are
\begin{equation}
    \label{SI-eq:adam-bias-correction}
    \hat M_t=\frac{M_t}{1-\beta_1^t},\qquad \hat V_t=\frac{V_t}{1-\beta_2^t}.
\end{equation}
The Adam update of \(W\) is 
\begin{equation}
    \label{SI-eq:adam-update}
    U_t=\Delta^{\rm Adam} W_t = -\eta_W\frac{\hat M_t}{\sqrt{\hat V_t}+\epsilon_{\rm Adam}},
\end{equation}
where the square root and division are applied entrywise. Thus Adam controls the entrywise scale of the update, but may keep nonzero mean in the columns.

To see the consequence, focus on \(\nabla_W^{(1)}\mcL_{\mathcal B}\). Using \(S=\tilde S+\mathbf 1_K\bar\sigma^\top\) from Eq. \eqref{SI-eq:S-mean-decomp}, we obtain
\begin{equation}
    \label{SI-eq:adam-grad1-decomp}
    \nabla_W^{(1)}\mcL_{\mathcal B}=\frac{s_2}{B}\tilde S R^\top+s_2\mathbf 1_K q^\top,\qquad q:=\frac{1}{B}R\bar\sigma .
\end{equation}
The first term corresponds to the bulk, whereas the second corresponds to the spike.

We approximate the Adam update, using Eqs. \eqref{SI-eq:adam-bias-correction} and \eqref{SI-eq:adam-update}, as
\begin{equation}
\label{eq:adam-sign-approx}
    U_{ki}\simeq -\eta_W \, \mathrm{sign} \left((\nabla_W\mcL_{\mathcal B})_{ki}\right).
\end{equation}
Keeping only the contribution from Eq. \eqref{SI-eq:adam-grad1-decomp}, write
\[
    \left(\nabla_W^{(1)}\mcL_{\mathcal B}\right)_{ki}=s_2(q_i+\xi_{ki}),\qquad \xi_{ki}=\frac{1}{B}\sum_{\mu=1}^B\tilde S_{k\mu}R_{i\mu}.
\]
Then
\begin{equation}
    U_{ki}\simeq -\eta_W\operatorname{sign}(q_i+\xi_{ki}).
    \label{SI-eq:adam-sign-q-xi}
\end{equation}
If \(q_i\neq0\), the signs in Eq. \eqref{SI-eq:adam-sign-q-xi} are biased in the same direction for many values of \(k\). Therefore the row average 
\begin{equation}
    \bar U_i=\frac{1}{K}\sum_{k=1}^KU_{ki}
    \label{SI-eq:adam-row-mean}
\end{equation}
is generically not \(O(\eta_W/\sqrt K)\), but can be \(O(\eta_W)\).

The output Adam update $\Delta F$ is obtained by replacing \( \Delta W \) with \( U \) in Eq. \eqref{SI-eq:dF-in-terms-dW} 
\begin{equation}
\label{eq:dF-adam-update}
    \Delta_{\rm Adam} F = s_2 U^\top S + s_1s_2W^\top\left(S'\odot(U G)\right)  := \Delta_{\rm Adam}^{(1)} F + \Delta_{\rm Adam}^{(2)} F . 
\end{equation}

\begin{figure}
    \centering
    \includegraphics[width=1\linewidth]{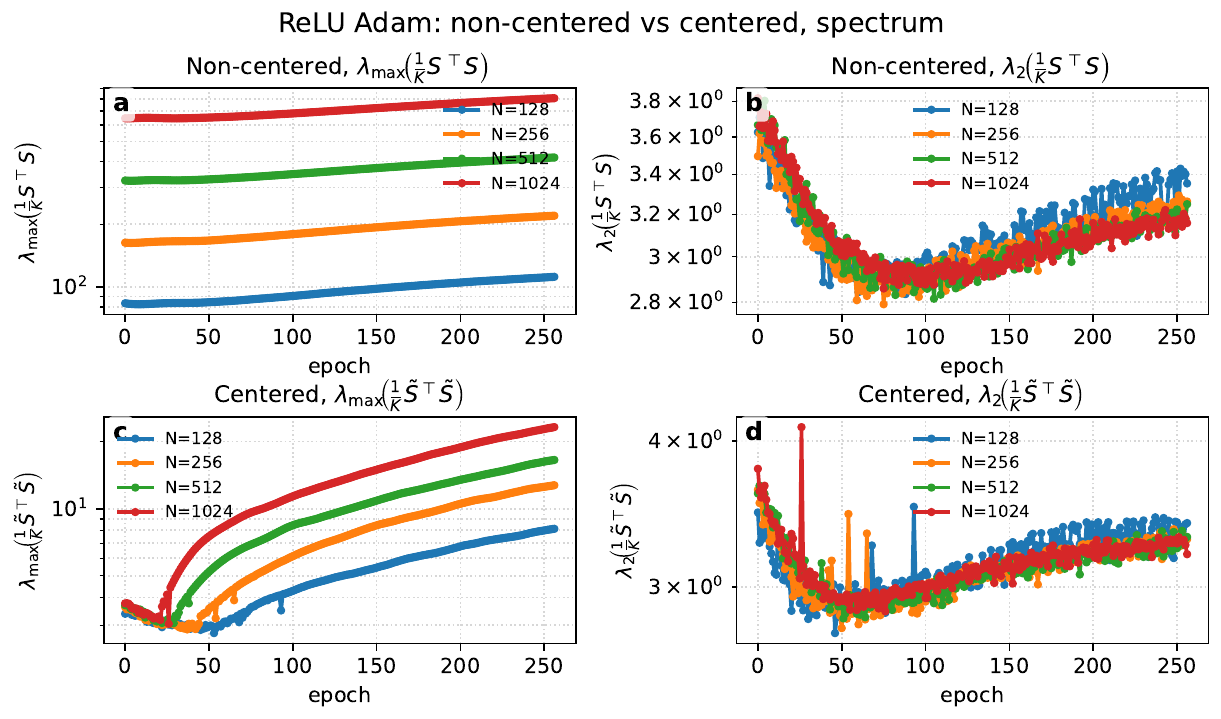}
    \caption{Spike and bulk behaviors for centered and non-centered ReLU DenseAM trained with mini-batch Adam in the proportional regime \eqref{eq:proportional-regime} with scale factors $\kappa=2, \rho=5, \beta=0.1$ and $256$ epochs. 
    In contrast to SGD, the Gram matrix $S^\top S$ does not enter directly into the updates.
    We show these diagnostics to provide a comparison with the SGD case in Fig. \ref{fig:center-non-center-relu-sgd-eigenvalues}.
    The left panels show the maximal eigenvalues $\lambda_{\max} (S^\top S / K)$ and $\lambda_{\max} (\tilde S^\top \tilde S / K)$, corresponding to the non-centered network and the centered network, respectively. 
    These reflect the spike discussed in App. \ref{SI-sec:adam}. 
    The right panels show the second largest eigenvalues $\lambda_{2} (S^\top S / K)$ and $\lambda_{2} (\tilde S^\top \tilde S / K)$, corresponding to the non-centered network and the centered network, respectively.
    These reflect the bulk eigenvalues.
    For Adam, centering does not yield collapse of the spike as seen in panels (b), but it weakens its increase (compare with \ref{fig:center-non-center-relu-sgd-eigenvalues}(c)).
    The appropriate observable for Adam is the actual update $U = \Delta^{\rm Adam} W$ in Eq. \eqref{SI-eq:adam-update}, plotted in Fig. \ref{fig:adam-update-center-non-center}.
    There we can see that centering achieves dynamical collapse.
    }
    \label{fig:center-non-center-relu-adam-eigenvalues}
\end{figure}

\begin{figure}
    \centering
    \includegraphics[width=1\linewidth]{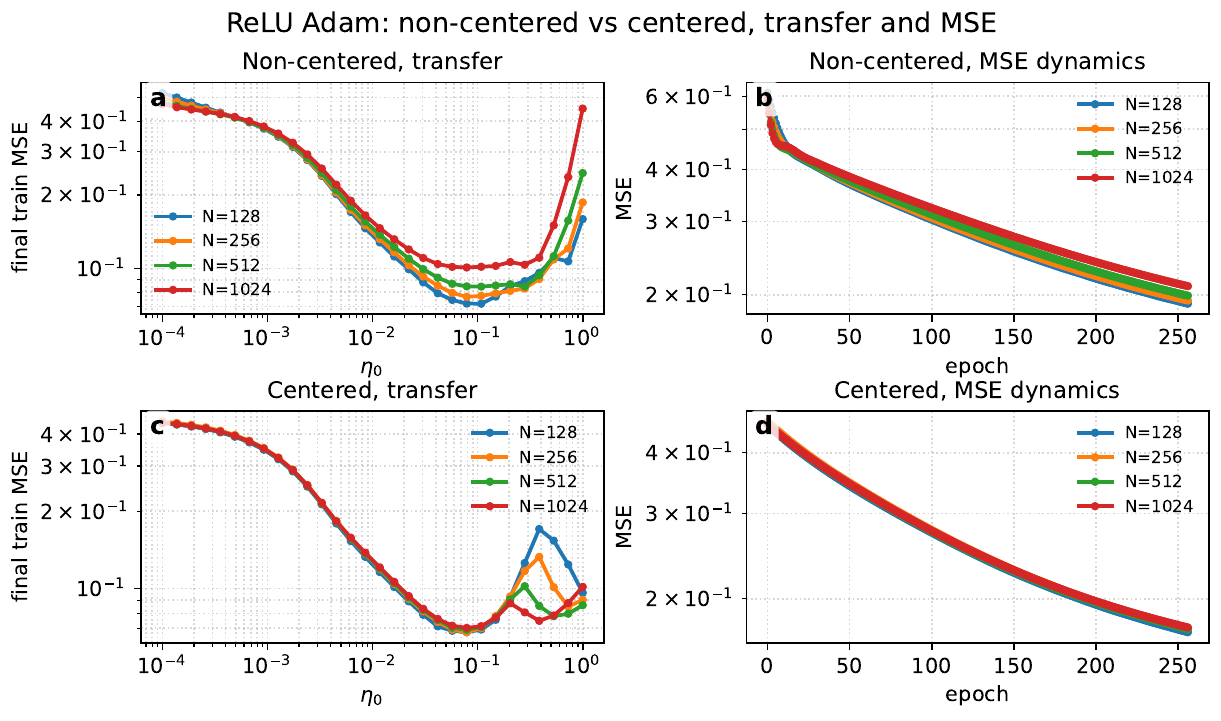}
    \caption{HP transfer for centered and non-centered ReLU DenseAM trained mini-batch Adam in the proportional regime \eqref{eq:proportional-regime} with scale factors $\kappa=2, \rho=5, \beta=0.1$ and $256$ epochs. 
    The left panels show the MSE loss after 256 training epochs as function of $\eta_0$ following the prescription in Table \ref{tab:results}, setting $s_1 =1/\sqrt{N}, s_2 =1/\sqrt{K}, \eta_W = \eta_0$. 
    Panel (c) showing the centered DenseAM exhibits better HP transfer than panel (a). 
    The right panels show the evolution of the MSE loss during training for $\eta_0 = 0.005$.
    The centered DenseAM (panel (d)) exhibits much better collapse across scales than the non-centered DenseAM (panel (b)).
    Compare with Fig. \ref{fig:center-non-center-relu-sgd-transfer} showing the same plot for SGD.
    See Fig. \ref{fig:adam-update-center-non-center} for the dynamics of the Adam update \eqref{SI-eq:adam-update}.
    }
    \label{fig:center-non-center-relu-adam-transfer}
\end{figure}

\begin{figure}
    \centering
    \includegraphics[width=1\linewidth]{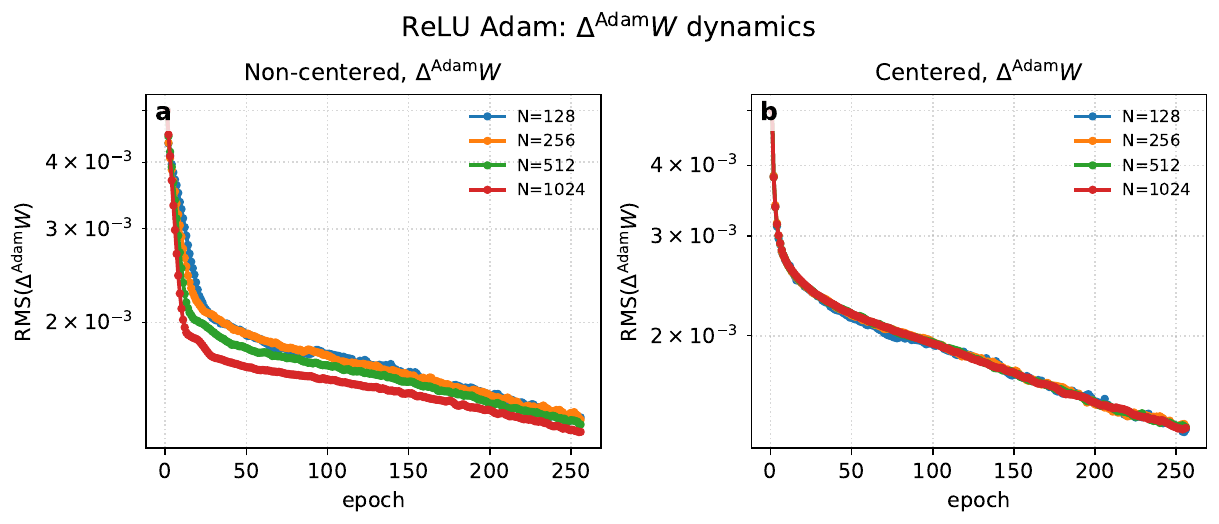}
    \caption{The Adam update \eqref{SI-eq:adam-update} for non-centered and centered ReLU DenseAM in the proportional regime, with $\kappa=2, \rho=5, \beta=0.1$ and $\eta_0 = 0.005$. Centering (b) provides good collapse of the update dynamics with respect to the non-centered DenseAM in (a).
    This corresponds to the better HP transfer in Fig. \ref{fig:center-non-center-relu-adam-transfer}(a,c).
    }
    \label{fig:adam-update-center-non-center}
\end{figure}

Now decompose the raw Adam update as
\begin{equation}
    \label{SI-eq:adam-U-decomp}
    U= \left( I_K - \frac{1}{K} \mathbf{1}_K \mathbf{1}_K^{\top} +  \frac{1}{K} \mathbf{1}_K \mathbf{1}_K^{\top}  \right) U =  CU+\mathbf 1_K\bar U^\top .
\end{equation}
Substitute this into the first term $\Delta_{\rm Adam}^{(1)} F$ in Eq. \eqref{eq:dF-adam-update}
\begin{align}
    \Delta_{\rm Adam}^{(1)}F &=s_2U^\top S =s_2(CU)^\top\tilde S+s_2(CU)^\top\mathbf 1_K\bar\sigma^\top+s_2\bar U\mathbf 1_K^\top\tilde S+s_2\bar U\mathbf 1_K^\top\mathbf 1_K\bar\sigma^\top \nonumber\\
    &=s_2(CU)^\top\tilde S+s_2K\bar U\bar\sigma^\top ,
    \label{SI-eq:adam-dF1-decomp}
\end{align}
where we used \((CU)^\top\mathbf 1_K=0\), \(\mathbf 1_K^\top\tilde S=0\), and \(\mathbf 1_K^\top\mathbf 1_K=K\).
Since the entries of \(CU\) are positive and negative terms of order \(O(\eta_W)\), it scales as
\begin{equation}
    s_2\left[(CU)^\top\tilde S\right]_{i\mu}\sim s_2\eta_W\sqrt K .
    \label{SI-eq:adam-fluctuating-scaling}
\end{equation}
However, the second term in Eq. \eqref{SI-eq:adam-dF1-decomp} gives a coherent contribution. If \(\bar U_i=O(\eta_W)\) and \(\bar\sigma_\mu=O(1)\), then
\begin{equation}
    s_2K\bar U_i\bar\sigma_\mu\sim s_2K\eta_W .
    \label{SI-eq:adam-rowmean-scaling}
\end{equation}
With the scaling \(s_2\sim K^{-1/2}\) from Eq. \eqref{SI-eq:s2-proportional-linear}, Eqs. \eqref{SI-eq:adam-fluctuating-scaling} and \eqref{SI-eq:adam-rowmean-scaling} yield different scaling with $K$
\begin{equation}
    \label{SI-eq:adam-two-scalings}
    s_2\eta_W\sqrt K \sim \eta_W,\qquad s_2K\eta_W \sim \eta_W\sqrt K.
\end{equation}
Thus even though Adam rescales the entrywise update, nonzero mean activations still lead to different scaling of $\Delta_{\rm Adam}^{(1)} F$, in the same way the spike and bulk appear for SGD in App. \ref{SI-sec:spike-nonzero-activation}.
Centering as in App. \ref{sec:centering} removes the spike and allows transfer. 
See Figs. \ref{fig:center-non-center-relu-adam-eigenvalues}, \ref{fig:center-non-center-relu-adam-transfer} for a comparison of centered and non-centered ReLU DenseAM trained with Adam.
In Fig. \ref{fig:adam-update-center-non-center} we show the scaling of the Adam update \eqref{SI-eq:adam-update} for non-centered and centered ReLU DenseAM.
Indeed, centering provides better dynamical collapse.

\subsection{Adam scaling for the centered model}
\label{sec:adam-for-centered-model}

For the centered model, the forward pass depends on \(W\) only through \(\tilde W=CW\). Therefore, if Adam applies the raw update \(W_{t+1} =  W_t+U_t\), the effective update seen by the network is
\begin{equation}
    \Delta\tilde W_t = C (W_t + U_t) - CW_t = CU_t .
    \label{SI-eq:centered-adam-effective-update}
\end{equation}
Thus the non-centered part of the raw Adam update is invisible to the network. We therefore apply the Adam scaling estimate to the effective update \(\Delta\tilde W=CU\).

As \(\Delta\tilde W=CU\), we have \((\Delta\tilde W)^\top\mathbf 1_K=0\). Therefore, decomposing \(S=\tilde S+\mathbf 1_K\bar\sigma^\top\),
\begin{equation}
    \label{SI-eq:centered-adam-removes-activation-mean}
    \Delta\tilde W^\top S = \Delta\tilde W^\top \left( \tilde S+\mathbf 1_K\bar\sigma^\top \right) = \Delta\tilde W^\top\tilde S .
\end{equation}
Thus, for the centered model, the nonzero mean of the activation does not contribute to the output-update term. Namely, Eq. \eqref{SI-eq:adam-dF1-decomp} becomes
\begin{equation}
    \Delta_{\rm Adam}^{(1)}F = s_2(CU)^\top\tilde S= s_2 U^\top \tilde S.
\end{equation}

Since \(U_{ki}=O(\eta_W)\), the pre-activations update is given by
\begin{equation}
    \label{SI-eq:centered-adam-dZ}
    \Delta_{\rm Adam} Z_{k\mu}=s_1\sum_{i=1}^N U_{ki} G_{i\mu}\sim \eta_Ws_1\sqrt N.
\end{equation}
The output update, Eq. \eqref{SI-eq:dF-in-terms-dW}, has the first-order contributions
\begin{align}
    \label{SI-eq:centered-adam-dF1}
  \left( \Delta^{(1)}_{\rm Adam} F \right)_{i \mu} &= s_2(U^\top \tilde S)_{i\mu}\sim s_2\eta_W\sqrt K  \\
    \label{SI-eq:centered-adam-dF2}
   \left( \Delta^{(2)}_{\rm Adam} F \right)_{i \mu} &=  s_1s_2\left[\tilde W^\top(S'\odot\Delta Z)\right]_{i\mu}\sim s_1s_2\eta_W\sqrt{KN}.
\end{align}
Thus the centered Adam prescription in the proportional regime \eqref{SI-eq:proportional-regime}, for elementwise activations is
\begin{equation}
    s_1\sim 1 / \sqrt{N},\qquad s_2 \sim 1 / \sqrt{K},\qquad \eta_W\sim 1 .
    \label{SI-eq:adam-centered-elementwise-prescription}
\end{equation}
See Figs. \ref{fig:center-non-center-relu-adam-eigenvalues}, \ref{fig:center-non-center-relu-adam-transfer} and \ref{fig:adam-update-center-non-center} for transfer and dynamics of ReLU DenseAM trained with Adam.

In Fig. \ref{fig:transfer-relu-p-large-p} we show HP transfer plots for various powers $p$ for ReLU$^p$ DenseAMs for both Adam and SGD.

\section{Bias learning rate scaling}
\label{SI-sec:bias}

The bias updates are structurally different from the weight updates because they create rank-one contributions across the sample dimension.
Below we provide heuristic arguments for the bias learning rates, which depend on whether sample averages are coherent or fluctuation-dominated.  

\subsection{Inner bias}

The hidden layer update, using \eqref{SI-eq:db}, is given by
\begin{equation}
\label{eq:SI-dZ_b}
    \Delta_b Z = -\eta_b\frac{s_2}{B} \left(S'\odot(WR)\right)\mathbf{1}_B\mathbf{1}_B^\top.
\end{equation}
For ease of notation, denote
\begin{equation}
    A=S'\odot(WR)\in\R^{K\times B}, \qquad
    \bar a=\frac{1}{B}A\mathbf{1}_B\in\R^K, \qquad \Delta_b Z = -\eta_b s_2\bar a\,\mathbf{1}_B^\top.
\end{equation}
Thus the Frobenius norm per entry is
\begin{equation}
\label{SI-eq:DbZ-frob}
    \frac{\norm{\Delta_b Z}_F^2}{KB} = \eta_b^2s_2^2\frac{\norm{\bar a}^2}{K}.
\end{equation}
Treating \(R\) as an order-one vector with no anomalous alignment with an individual row of \(W\),
\[
    (WR)_{k\mu}=O(\sqrt N),
\]
and hence \(A_{k\mu}=O(\sqrt N)\).  There are then two natural possibilities for the sample average:
\begin{equation}
    \bar a_k = \frac{1}{B} \sum_{\mu=1}^B A_{k \mu} \sim
    \begin{cases}
        \sqrt N, & \text{coherent average},\\
        \sqrt{N/B}, & \text{fluctuation average}.
    \end{cases}
\end{equation}
Therefore the norm scales as
\begin{equation}
\label{SI-eq:abar-norm}
    \norm{\bar a}^2 \sim
    \begin{cases}
        KN, & \text{coherent average},\\
        KN/B, & \text{fluctuation average}.
    \end{cases}
\end{equation}
Using \(s_2\sim K^{-1/2}\), plugging \eqref{SI-eq:abar-norm} into \eqref{SI-eq:DbZ-frob} we get
\begin{equation}
\label{SI-eq:etab-cases}
    \frac{\norm{\Delta_b Z}_F^2}{KB} \sim \eta_b^2
    \begin{cases}
        N/K, & \text{coherent average},\\
        N/(KB), & \text{fluctuation average}.
    \end{cases}
\end{equation}
In the proportional regime, the coherent case gives the prescription
\begin{equation}
    \eta_b=O(1),
\end{equation}
while the fluctuation case would allow \(\eta_b=O(\sqrt N)\).
Empirically, the \(O(1)\) prescription is more stable and is the prescription used in Table \ref{tab:results}.

\subsection{Outer bias}

The output update due to a change of the outer bias $c$ is given by (recall \eqref{SI-eq:dc}),
\begin{equation}
    \Delta_cF = -\eta_c\bar r\,\mathbf{1}_B^\top, \qquad \bar r=\frac{1}{B}R\mathbf{1}_B.
\end{equation}
Hence, the Frobenius norm per entry is
\begin{equation}
\label{SI-eq:DcF-frob}
    \frac{\norm{\Delta_cF}_F^2}{NB} = \eta_c^2\frac{\norm{\bar r}^2}{N}.
\end{equation}
If \(R_{i\mu}=O(1)\), then
\begin{equation}
    \bar r_i = \frac{1}{B} \sum_{\mu=1}^B R_{i \mu} \sim
    \begin{cases}
        1, & \text{coherent average},\\
        1 / \sqrt{B}, & \text{fluctuation average}.
    \end{cases}
\end{equation}
Thus, plugging this into \eqref{SI-eq:DcF-frob}
\begin{equation}
\label{SI-eq:etac-cases}
    \frac{\norm{\Delta_cF}_F^2}{NB} \sim \eta_c^2
    \begin{cases}
        1, & \text{coherent average},\\[3pt]
        1/B, & \text{fluctuation average}.
    \end{cases}
\end{equation}
The coherent scaling gives
\begin{equation}
    \eta_c=O(1),
\end{equation}
while the fluctuation scaling would allow \(\eta_c=O(\sqrt B)\).  As for the inner bias, we use  \(\eta_c = O(1)\) as the  prescription in Table \ref{tab:results}.

\section{Softmax activations}
\label{SI-sec:softmax}

Softmax must be treated separately because it is not an elementwise activation, and therefore its Jacobian is not diagonal.
Softmax is given by
\begin{equation}
\label{SI-eq:softmax-def}
    \sigma_k=(\sigma(z))_k=\frac{e^{z_k}}{\sum_{j=1}^K e^{z_j}},
\end{equation}
and its Jacobian is
\begin{equation}
\label{SI-eq:softmax-jacobian}
    J=\frac{\partial \sigma}{\partial z}=\operatorname{diag}(\sigma)-\sigma\sigma^\top.
\end{equation}
As the inner bias $b$ does not affect the softmax activation, we disregard it in the following analysis.

\subsection{Initialization scales for softmax}
\label{SI-sec:softmax-initialization}

We first fix the initialization scales.
Recall the centered model \eqref{eq:centered-model},
\[
    f=s_2\tilde W^\top\sigma(z)+c,\qquad z=s_1\tilde W g,  \qquad \tilde W=CW.
\]
As in Sec.~\ref{SI-sec:s1-scaling}, the logits are sums of \(N\) order-one terms, so \(z_k=O(1)\) requires
\begin{equation}
\label{SI-eq:softmax-s1}
    s_1\sim \frac{1}{\sqrt N}.
\end{equation}
For \(O(1)\) logits, we assume the softmax is delocalized at initialization, meaning
\begin{equation}
\label{SI-eq:softmax-delocalized}
    \|\sigma\|^2=\sum_{k=1}^K \sigma_k^2=O\left(\frac{1}{K}\right).
\end{equation}
Since \(\sum_k\sigma_k=1\), we have the exact identity
\[
    \|C\sigma\|^2=\|\sigma\|^2-\frac{1}{K}.
\]
Thus if the softmax is delocalized but not asymptotically uniform, then \(\|C\sigma\|^2=\Theta(1/K)\).
In this case, centering removes the uniform component but does not change the \(K\)-scaling of the norm.
If the logits are nearly equal, this centered norm can be smaller, and it vanishes for a perfectly uniform softmax.

Next, we treat the scale \( s_2 \) by considering the output scale \( f_i \).
Let \(u=C\sigma(z)\), with \(z=s_1CWg\). Then the shifted softmax output (i.e., removing the bias \( c \)) satisfies
\begin{equation}
    \label{eq:softmax-f-scaling}
    f_i-c_i = s_2 \left( W^\top u\right)_i = s_2  \sum_{k=1}^K W_{ki}u_k .
\end{equation}
By Gaussian integration by parts,
\[
    \mathbb E_W\left[\left(W^\top u\right)_i\,\middle|\,g\right]=\sum_{k=1}^K \mathbb E_W\left[W_{ki}u_k\,\middle|\,g\right]=\sum_{k=1}^K \mathbb E_W\left[\frac{\partial u_k}{\partial W_{ki}}\,\middle|\,g\right].
\]
Since \(u=C\sigma(z)\) and \(z=s_1CWg\),
\[
    \frac{\partial u}{\partial W_{ki}}=s_1g_iCJC e_k.
\]
Therefore, Eq. \eqref{eq:softmax-f-scaling} becomes
\begin{equation}
    \label{eq:softmax-f-scaling-2}
     f_i-c_i = s_2\sum_{k=1}^K\frac{\partial u_k}{\partial W_{ki}} =s_2 s_1 g_i\operatorname{Tr}(CJC).
\end{equation}
The softmax Jacobian, Eq. \eqref{SI-eq:softmax-jacobian}, satisfies \(CJ=JC=J\), hence
\[
    \operatorname{Tr}(CJC)=\operatorname{Tr}J=\operatorname{Tr}\left(\operatorname{diag}(\sigma)-\sigma\sigma^\top\right)=1-\|\sigma\|^2.
\]
Thus, Eq. \eqref{eq:softmax-f-scaling-2} becomes
\[
    \mathbb E_W\left[f_i-c_i\,\middle|\,g\right]=s_2s_1g_i\,\mathbb E_W\left[1-\|\sigma\|^2\,\middle|\,g\right].
\]
For a delocalized softmax, \(\|\sigma\|^2=O(1/K)\), so \(1-\|\sigma\|^2=O(1)\). With \(g_i=O(1)\), this gives
\[
    \mathbb E_W\left[f_i-c_i\,\middle|\,g\right]=O(s_2s_1).
\]
Therefore keeping this contribution order one requires \(s_2s_1=O(1)\). Using \(s_1\sim N^{-1/2}\) in Eq. \eqref{SI-eq:softmax-s1}, we obtain \(s_2\sim\sqrt N\). In the proportional ratio regime \(K=\kappa N\), and therefore we set
\begin{equation}
\label{SI-eq:softmax-s2}
    s_2\sim \sqrt K .
\end{equation}
Note that difference from the elementwise activations where we got $s_2 \sim 1 / \sqrt{K}$, see Table \ref{tab:results}.

\subsection{Softmax spike and centering}

For every sample, softmax is normalized, \(   \sum_{k=1}^K \sigma_k^\mu=1 \), and so, its mean is exactly \(  \bar\sigma_\mu= 1 / K \).
Decomposing \(S\) into (as in Eq. \eqref{SI-eq:sigma-bar-def})
\begin{equation}
\label{SI-eq:softmax-S-decomp}
    S=\tilde S+\frac{1}{K}\mathbf{1}_K\mathbf{1}_B^\top,
\end{equation}
we get
\begin{equation}
\label{SI-eq:softmax-gram-decomp}
    S^\top S = \tilde S^\top\tilde S + \frac{1}{K}\mathbf{1}_B\mathbf{1}_B^\top.
\end{equation}
Compare with the elementwise activations in Eqs. \eqref{SI-eq:S-mean-decomp} and \eqref{SI-eq:S-gram-decomp}.
The spike eigenvalue is (compare with Eq. \eqref{SI-eq:lambda-spike-elementwise})
\begin{equation}
\label{SI-eq:softmax-spike}
    \lambda_{\rm spike}(S^\top S) = \frac{B}{K}.
\end{equation}
In the proportional regime $B \sim K$ and so \(\lambda_{\rm spike}(S^\top S) = O(1)\).
If the softmax is delocalized, \(\sigma_k^\mu=O(1/K)\), the centered entries are also \(O(1/K)\), so the bulk eigenvalues of \(\tilde S^\top\tilde S\) are \(O(1/K)\).
Thus the mean spike dominates the delocalized softmax bulk and should be removed by centering for stability.

\begin{figure}
    \centering
    \includegraphics[width=1\linewidth]{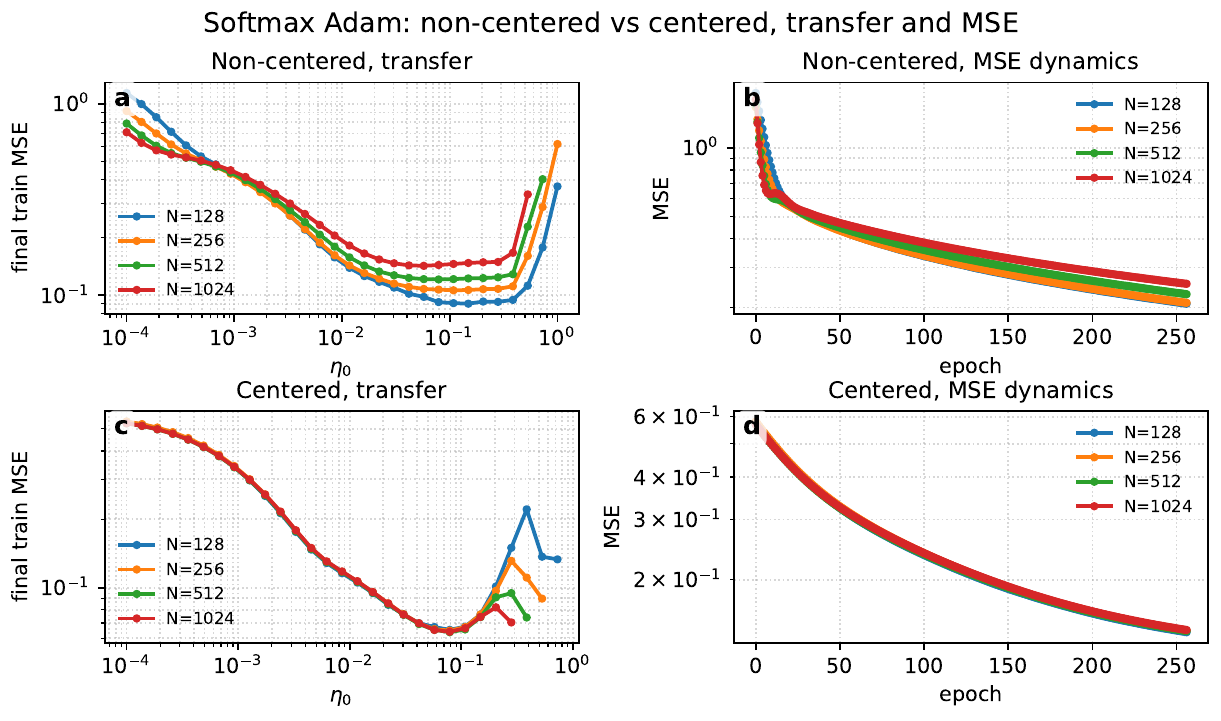}
    \caption{LR transfer for DenseAM with centered vs. non-centered softmax activations trained with Adam in the proportional regime \eqref{eq:proportional-regime} with $\kappa=2, \rho=5, \beta=0.1$ trained for 256 epochs and $s_1 =1/\sqrt{N}, s_2 = \sqrt{K}, \eta_W = \eta_0$. 
    Panels (a,c) show the final MSE loss as a function of $\eta_0$.
    We observe improved HP transfer by centering.
    Panels (b,d) show the training dynamics of the loss for $\eta_0=0.005$.
    Centering exhibits better collapse as we scale up the DenseAM.
    }
    \label{fig:softmax-adam-center-non-center-transfer-and-mse}
\end{figure}

In Fig. \ref{fig:softmax-adam-center-non-center-transfer-and-mse} we compare between non-centered and centered softmax DenseAM for Adam training.
We observe that centering is indeed necessary for dynamical collapse and for HP transfer.

\subsection{Row amplification feedback makes SGD unstable}

Softmax has an additional instability mechanism that is not removed by centering.
Let \(w_k\in\R^N\) denote the \(k\)-th row of \(W\).
We start with the uncentered model \eqref{eq:SI-uncentered-model} written in terms of rows of \( W \)
\[
    f=s_2\sum_{i=1}^K w_i \sigma_i.
\]
Varying only row \(w_k\), we have
\begin{equation}
\label{SI-eq:softmax-deltaf-row}
    \delta f = s_2\sigma_k\,\delta w_k + s_1s_2\sum_{i=1}^K w_i J_{ik}(\delta w_k^\top g).
\end{equation}
For the single-sample loss
\[
    \ell=\frac{1}{2}\norm{r}^2, \qquad r=f-y,
\]
we get
\begin{align}
\label{SI-eq:softmax-row-gradient}
    \frac{\partial \ell}{\partial w_k} &= s_2\sigma_k r + s_1s_2  \left(JWr\right)_k g = s_2\sigma_k \left[ r + s_1\left((Wr)_k-\sigma^\top Wr\right)g \right].
\end{align}
Thus the row gradient is proportional to the current softmax weight \(\sigma_k\).  
This means that rows with larger softmax weights receive larger SGD updates, which can amplify softmax concentration and spoil transfer.

\begin{figure}
    \centering
    \includegraphics[width=1\linewidth]{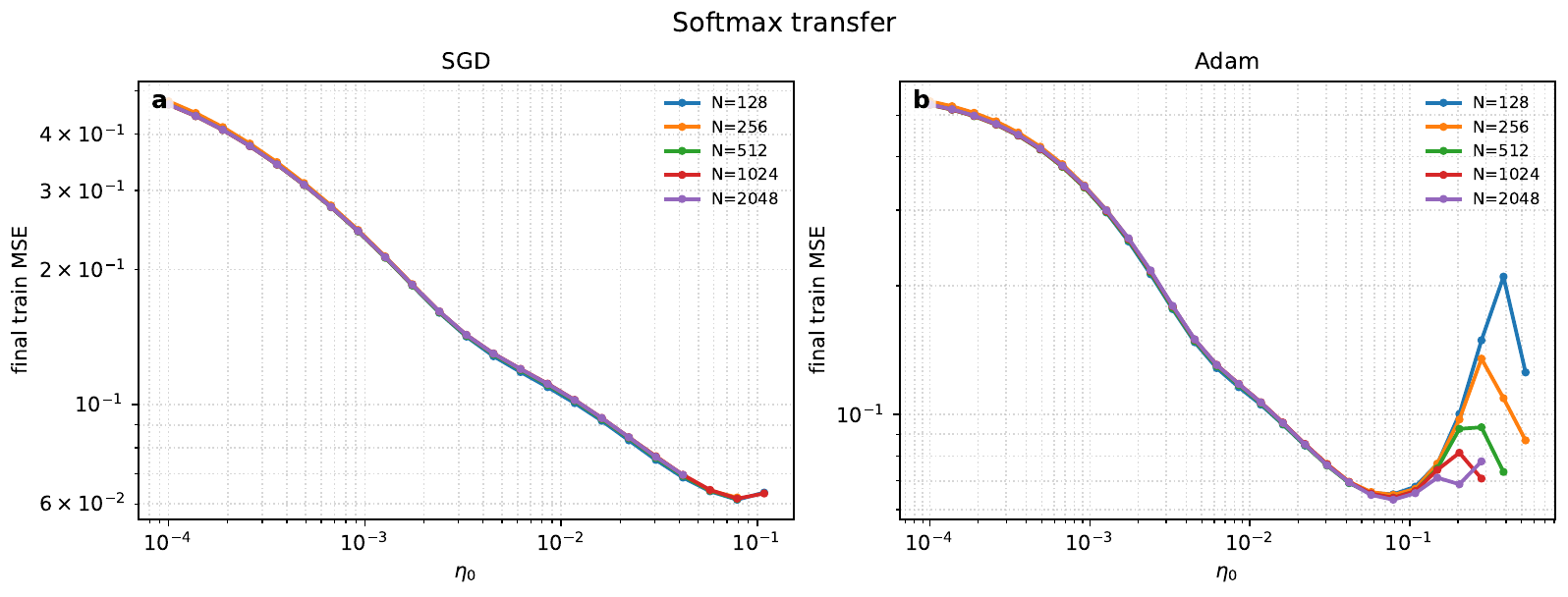}
    \caption{
    Learning rate transfer for softmax DenseAM trained with SGD and Adam in the proportional regime \eqref{eq:proportional-regime} with $\kappa=2, \rho=5, \beta=0.1$ trained for 256 epochs. 
    For both optimizers we use $s_1 =1/\sqrt{N}, s_2 = \sqrt{K}$, but the learning rate differs: For SGD we use $ \eta_W = \eta_0 K$ whereas for Adam we use $\eta_W = \eta_0$ (as in Table \ref{tab:results}).
    For SGD training (a) we observe instability at $N=2048$ as the optimal learning rate is smaller than for other values of $N$ (see similar but more drastic instance of this in Fig. \ref{fig:center-non-center-relu-sgd-transfer}(a)), whereas Adam (b) does not exhibit this instability. This is further illustrated in Fig. \ref{fig:softmax-dynamics-isntability} where the training dynamics exhibits that for SGD the localization of the softmax activation is correlated with the increase of the weight updates during training. 
    This instability does not appear for Adam.}
    \label{fig:transfer-softmax-sgd-vs-adam}
\end{figure}

\begin{figure}
    \centering
    \includegraphics[width=1\linewidth]{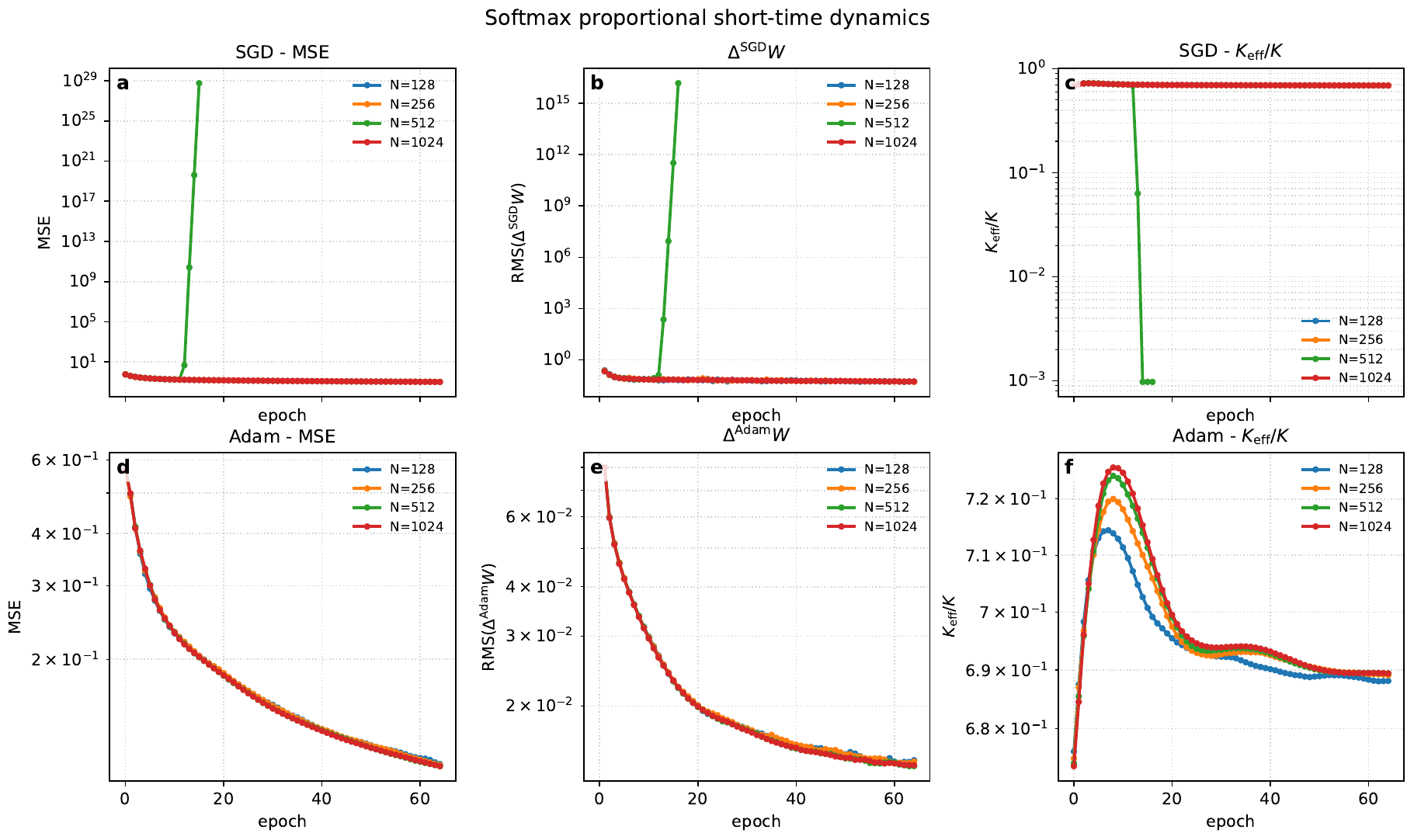}
    \caption{
    Training dynamics of softmax DenseAM with Adam in the proportional regime \eqref{eq:proportional-regime} with $\kappa=2, \rho=5, \beta=0.1$ with $\eta_0 = 0.08$. 
    For both optimizers we use $s_1 =1/\sqrt{N}, s_2 = \sqrt{K}$, but the learning rate differs: For SGD we use $ \eta_W = \eta_0 K$ whereas for Adam we use $\eta_W = \eta_0$ (as in Table \ref{tab:results}).
    Here, $\Delta^{\rm Adam} W$ \eqref{SI-eq:adam-update} it the update induced by Adam, and $K_{\rm eff} = 1 / \sum_{k=1} ^K\sigma_k^2$ is the softmax activation participation ratio \eqref{eq:K-eff-def} which measures how delocalized is softmax: $K_{\rm eff} \approx K$ means the softmax is delocalized, whereas $K_{\rm eff} \approx 1$ means the softmax is localized.
    SGD (top row) can lead to a rapid decrease of the softmax participation ratio, reflecting the localization of softmax, leading to numerical instability. Adam (bottom row), in contrast, avoids this effect. Thus we observe HP transfer only for Adam with softmax, see Fig. \ref{fig:transfer-softmax-sgd-vs-adam}.}
    \label{fig:softmax-dynamics-isntability}
\end{figure}

For the centered model \eqref{eq:centered-model}, the same calculation gives
\[
    \frac{\partial \ell}{\partial w_k}= s_2\left(\sigma_k-\frac{1}{K}\right)r + s_1s_2\sigma_k\left((Wr)_k-\sigma^\top Wr\right)g.
\]
Thus centering removes the uniform part of the direct readout update, replacing \(\sigma_k\) by \(\sigma_k-1/K\).
However, the other term still carries a factor \(\sigma_k\).
Therefore rows with anomalously large softmax weights can still receive anomalously large updates, which can amplify concentration. This is distinct from the mean spike in \(S^\top S\), which is removed by centering.
See Fig. \ref{fig:transfer-softmax-sgd-vs-adam}(a) for HP transfer instability of SGD, and Fig. \ref{fig:softmax-dynamics-isntability} for the dynamical instability as it occurs during training.
Note that, due to this instability, we have not specified a prescription for scaling in the case of softmax DenseAM with SGD in Table \ref{tab:results}. 
In the expriments shown in the Figures \ref{fig:transfer-softmax-sgd-vs-adam}, \ref{fig:softmax-dynamics-isntability}, for SGD we use $s_1 =1/\sqrt{N}, s_2 = \sqrt{K}, \eta_W = \eta_0 K$, whereas for Adam we use $s_1 =1/\sqrt{N}, s_2 = \sqrt{K}, \eta_W = \eta_0$.

\subsection{Adam scaling in the delocalized softmax regime}
\label{SI-sec:softmax-adam-scaling}

After concluding that softmax DenseAMs are unstable with SGD, we next provide a heuristic argument for the appropriate scaling in the proportional regime for Adam.
We use the same approximation for the Adam update as in \eqref{eq:adam-sign-approx} in App. \ref{SI-sec:adam-spike}
\begin{equation}
    \label{SI-eq:softmax-adam-U-sign}
    U_{ki}=\Delta W^{\rm Adam}_{ki}\approx -\eta_W\operatorname{sign}\left((\nabla_W\mcL_{\mathcal B})_{ki}\right).
\end{equation}
Therefore, the magnitude of the Adam updates are \(U_{ki}=O(\eta_W)\). 
For the centered model, the forward pass depends on \(W\) through \(\tilde W=CW\), so the effective update is \(CU\) (recall the discussion in App. \ref{sec:adam-for-centered-model}). 
The preactivation update (in vector form) is therefore
\begin{equation}
    \label{SI-eq:softmax-adam-dz}
    \Delta z=s_1CUg,\qquad \Delta z_k=s_1\left[(Ug)_k-\frac{1}{K}\sum_{\ell=1}^K(Ug)_\ell\right].
\end{equation}
Assuming random-sign summation over the input index, \((Ug)_k\sim \eta_W\sqrt N\). The projection by \(C\) does not increase the scale, so
\begin{equation}
    \label{SI-eq:softmax-adam-dz-scaling}
    \Delta z_k\sim \eta_Ws_1\sqrt N\sim \eta_W,
\end{equation}
where we used \(s_1\sim 1/\sqrt N\) from Eq.~\eqref{SI-eq:softmax-s1}. Thus \(\eta_W=O(1)\) gives order-one logit updates.

We next estimate the output update. For the centered softmax model \eqref{eq:centered-model}, the first-order update is
\begin{equation}
    \label{SI-eq:softmax-adam-df}
    \Delta f=s_2U^\top C\sigma+s_2W^\top CJ\Delta z ,  \qquad  J=\operatorname{diag}(\sigma)-\sigma\sigma^\top .
\end{equation}
Here the Adam update \(U\), Eq. \eqref{SI-eq:adam-update}, replaces \(\Delta W\), and the softmax Jacobian \(J\), Eq. \eqref{SI-eq:softmax-jacobian}, replaces the elementwise expression involving \(S'\odot\Delta Z\) in Eq.~\eqref{SI-eq:dF-in-terms-dW}. Since \(J\mathbf 1_K = \mathbf 1_K^\top J =0\), we have \(CJ=J\). Thus Eq.~\eqref{SI-eq:softmax-adam-df} becomes
\begin{equation}
    \label{SI-eq:softmax-adam-df-simplified}
    \Delta f= s_2 U^\top C\sigma + s_2 W^\top J\Delta z := \Delta^{(1)}_{\rm softmax} f +  \Delta^{(2)}_{\rm softmax} f .
\end{equation}

Next, we analyze the magnitude of these two terms in \eqref{SI-eq:softmax-adam-df-simplified}.
The first term \(\Delta^{(1)}_{\rm softmax} f =  s_2 U^\top C\sigma\) is a sum of $K$ terms. 
However, since $\sigma$ is normalized, the number of its nonzero coordinates is captured by the participation ratio, which we denote as the effective width
\begin{equation}
    \label{eq:K-eff-def}
    K_{\rm eff}:=\frac{1}{\sum_{k=1}^K\sigma_k^2}.
\end{equation}
If softmax is delocalized, i.e., many coordinates $\sigma_k$ are nonzero, then $K_{\rm eff} \approx K$.
However, when only a few coordinates $\sigma_k$ are nonzero, then $K_{\rm eff}$ becomes order one.

To estimate the scaling of the first term \(\Delta^{(1)}_{\rm softmax} f =  s_2 U^\top C\sigma\), note that the relevant object is $ C \sigma$ rather than $\sigma$.
Therefore, it is useful to consider an effective centered width defined as 
\begin{equation}
    \label{SI-eq:softmax-centered-Keff}
    \tilde K_{\rm eff}:=\frac{1}{\|C\sigma\|^2} =\frac{1}{K_{\rm eff}^{-1}-K^{-1}}  = \frac{K K_{\rm eff}}{K - K_{\rm eff}}.
\end{equation}
If the softmax is delocalized but not perfectly uniform, then \(\|C\sigma\| \sim K^{-1/2}\) and \(\tilde K_{\rm eff} \sim K\). If the softmax is nearly uniform, then \(C\sigma\) is smaller and \(\tilde K_{\rm eff}\gg K\). Finally, if the softmax is localized on \(O(1)\) coordinates, then \(\tilde K_{\rm eff}\sim 1\).
In the following, we assume the softmax remains in the delocalized but not perfectly uniform regime, i.e., \(\tilde K_{\rm eff} \sim K_{\rm eff} \sim K \).
This is supported by the dynamics during training shown in Fig. \ref{fig:softmax-dynamics-isntability}(f).

Now, we estimate the scaling of the first term \(\Delta^{(1)}_{\rm softmax} f =  s_2 U^\top C\sigma\).
If there is no coherent alignment between the Adam update $U$ and $C \sigma$, then, as $U \sim \eta_W$ and $\norm{C \sigma_k} \sim 1/ \tilde K_{\rm eff}^{-1/2}$, we get
\begin{equation}
    \label{SI-eq:softmax-adam-df-term1-Keff}
    \Delta^{(1)}_{\rm softmax} f = s_2 U^\top C \sigma \sim s_2\eta_W \norm{C\sigma}  \sim s_2\eta_W \tilde K_{\rm eff}^{-1/2} \sim s_2 \eta_W K_{\rm eff}^{-1 /2}.
\end{equation}

For the second term $\Delta^{(2)}_{\rm softmax} f $ in Eq. \eqref{SI-eq:softmax-adam-df-simplified}, let us consider first
\begin{equation}
    \label{SI-eq:softmax-adam-Jdz-entry-exact}
    (J\Delta z)_k=\sigma_k\left(\Delta z_k-\sum_j\sigma_j\Delta z_j\right).
\end{equation}
In the delocalized regime, \(\sigma_k \sim 1/K_{\rm eff}\), while \(\Delta z_k \sim \eta_W \) (recall Eq. \eqref{SI-eq:softmax-adam-dz-scaling}). The weighted average \(\sum_j\sigma_j\Delta z_j\) is also \(O(\eta_W)\), even if \(\Delta z\) is correlated with \(\sigma\). Therefore
\begin{equation}
    \label{SI-eq:softmax-adam-Jdz-scaling}
    (J\Delta z)_k \sim \eta_W/K_{\rm eff}.
\end{equation}
Next, \(W^\top J\Delta z\) is a sum of $K$ terms, out of which $K_{\rm eff}$ are nonzero. 
We further assume that \( J \Delta z\) is not coherently aligned with the columns of \(W\), so the sum over \(k\) has random-sign terms, and therefore
\begin{equation}
    \label{SI-eq:softmax-adam-df-term2}
   \Delta^{(2)}_{\rm softmax} f  = s_2(W^\top J\Delta z)_i\sim s_2\eta_W  / \sqrt{K_{\rm eff}}.
\end{equation}
If such alignment developed during training, \(\Delta^{(2)}_{\rm softmax} f\) could be \(O(s_2\eta_W)\) rather than \(O(s_2 \eta_W /\sqrt{K_{\rm eff}})\).

We find that the two terms \(\Delta^{(1)}_{\rm softmax} f, \, \Delta^{(2)}_{\rm softmax} f\) (Eqs. \eqref{SI-eq:softmax-adam-df-term1-Keff}, \eqref{SI-eq:softmax-adam-df-term2}), both scale the same, 
\begin{equation}
\label{SI-eq:softmax-adam-df-scaling}
    \Delta^{(1)}_{\rm softmax} f \sim \, \Delta^{(2)}_{\rm softmax} f \sim  s_2\eta_W  / \sqrt{K_{\rm eff}} \sim \eta_W \sqrt{K / K_{\rm eff}} 
\end{equation}
where we used $s_2 \sim \sqrt{K}$ from \eqref{SI-eq:softmax-s2}.

As the preactivations update $\Delta z$ in \eqref{SI-eq:softmax-adam-dz-scaling} requires $\eta_W \sim 1$, we find that the output update \eqref{SI-eq:softmax-adam-df-scaling} requires that we remain in the delocalized regime $K_{\rm eff} \sim K$.
If $K_{\rm eff}$ becomes to small, the output update becomes too large, and we expect no HP transfer.
Our numerical experiments suggest that we remain in this regime during training, see bottom panels of Fig. \ref{fig:softmax-dynamics-isntability}.
Moreover,  we observe that for SGD, the drop in $K_{\rm eff}$, i.e. the localization of softmax, is correlated with the divergence of the training updates, supporting this analysis. See top panels of Fig. \ref{fig:softmax-dynamics-isntability}.

\section{$K$-only scaling regime}
\label{sec:K-only-scaling}

In this section we analyze the $K$-only scaling, namely, $K \to \infty$, while $N, P, B$ are fixed.
To keep the analysis simple we will consider centering the network only when it is relevant.

\subsection{Elementwise activations}

We start by considering elementwise activations such as $\mathrm{ReLU}^p$.

\subsubsection{Preactivations and outputs scaling}

The initialization analysis of the hidden preactivations and outputs in App. \ref{SI-sec:initialization} is applicable here.
Keeping elements of $Z$ at initialization order one, requires $s_1 \sim 1 / \sqrt{N}$, see Eq. \eqref{SI-eq:s1-scaling}.
Note that $N$ is fixed, so this is just a number.
As for the outputs, we already found that $s_2 \sim 1 / K$ in this regime in Eq. \eqref{SI-eq:s2-onlyK-linear}.
Therefore the scaling factors are
\begin{equation}
\label{eq:K-only-s1-s2}
    s_1 \sim 1/\sqrt{N}, \qquad s_2 \sim 1 / K, \qquad \text{$K$-only regime}.
\end{equation}

\subsubsection{SGD updates of preactivations and outpus scaling}

Next, we consider the hidden preactivations update $\Delta_W Z$ in Eq. \eqref{SI-eq:dZ-expanded}.
Consider the first term 
\begin{equation}
\label{eq:dZ1-only-k-1st-step}
    \Delta_W^{(1)} Z = -\eta_W\frac{s_1 s_2}{B} S R^\top G.
\end{equation}
Recall that the dimensions of the matrices in this expressions are $S \in \R^{K \times B}, R \in \R^{N \times B}, G \in \R^{N \times B}$.
Therefore, there is no sum over $K$, and hence each entry scales as (using $s_2 \sim 1 / K$ from \eqref{eq:K-only-s1-s2})
\begin{equation}
\label{eq:dZ1-only-K-regime}
    \left( \Delta_W^{(1)} Z \right)_{i \mu} \sim \eta_W s_1 s_2 \sim \eta_W / K.
\end{equation}
The same applies for $\Delta_W^{(2)} Z$.

Let us analyze the output updates $\Delta_W F$ in Eq. \eqref{SI-eq:dF-decomp}.
Consider the centered version of the first output update term (Eq. \eqref{SI-eq:dF-11})
\begin{equation}
\label{eq:dF-11-only-K-regime}
\Delta_{\tilde W}^{(1,1)} F = -\eta_W\frac{s_2^2}{B}R \tilde S^\top \tilde S.
\end{equation}
The $K$ dependence appears only in the centered Gram matrix \( \tilde S^\top \tilde S / K\).
By the law of large numbers
\[
\lim_{K \to \infty} \frac{1}{K} \tilde S^\top \tilde S  = \Sigma_\sigma, \qquad \left(\Sigma_\sigma \right)_{\mu \nu} = {\rm Cov} \left( \sigma^\mu, \sigma^\nu \right).
\]
Therefore Eq. \eqref{eq:dF-11-only-K-regime} scales as (using $s_2 \sim 1 /K$ from \eqref{eq:K-only-s1-s2})
\begin{equation}
\label{eq:dF-11-only-K-scaling}
\Delta_{\tilde W}^{(1,1)} F  \sim \eta_W K \frac{s_2^2}{B} R \Sigma_\sigma \sim \eta_W K s_2^2 \sim \eta_W / K.
\end{equation}
The same scaling applies to the other terms in Eq. \eqref{SI-eq:dF-4-terms} that compose $\Delta_W F$.

Finally, to keep both updates $\Delta_W Z$ (Eq. \eqref{eq:dZ1-only-K-regime}) and $\Delta_W F$ (Eq. \eqref{eq:dF-11-only-K-scaling}) order one, we ought to set $\eta_W \sim K$, and hence we get the prescription
\begin{equation}
\label{eq:only-K-scaling-final}
        s_1 \sim 1/\sqrt{N}, \quad s_2 \sim 1 / K, \quad \eta_W \sim K, \qquad \text{SGD, elementwise activations, $K$-only regime}.
\end{equation}

\begin{figure}
    \centering
    \includegraphics[width=1\linewidth]{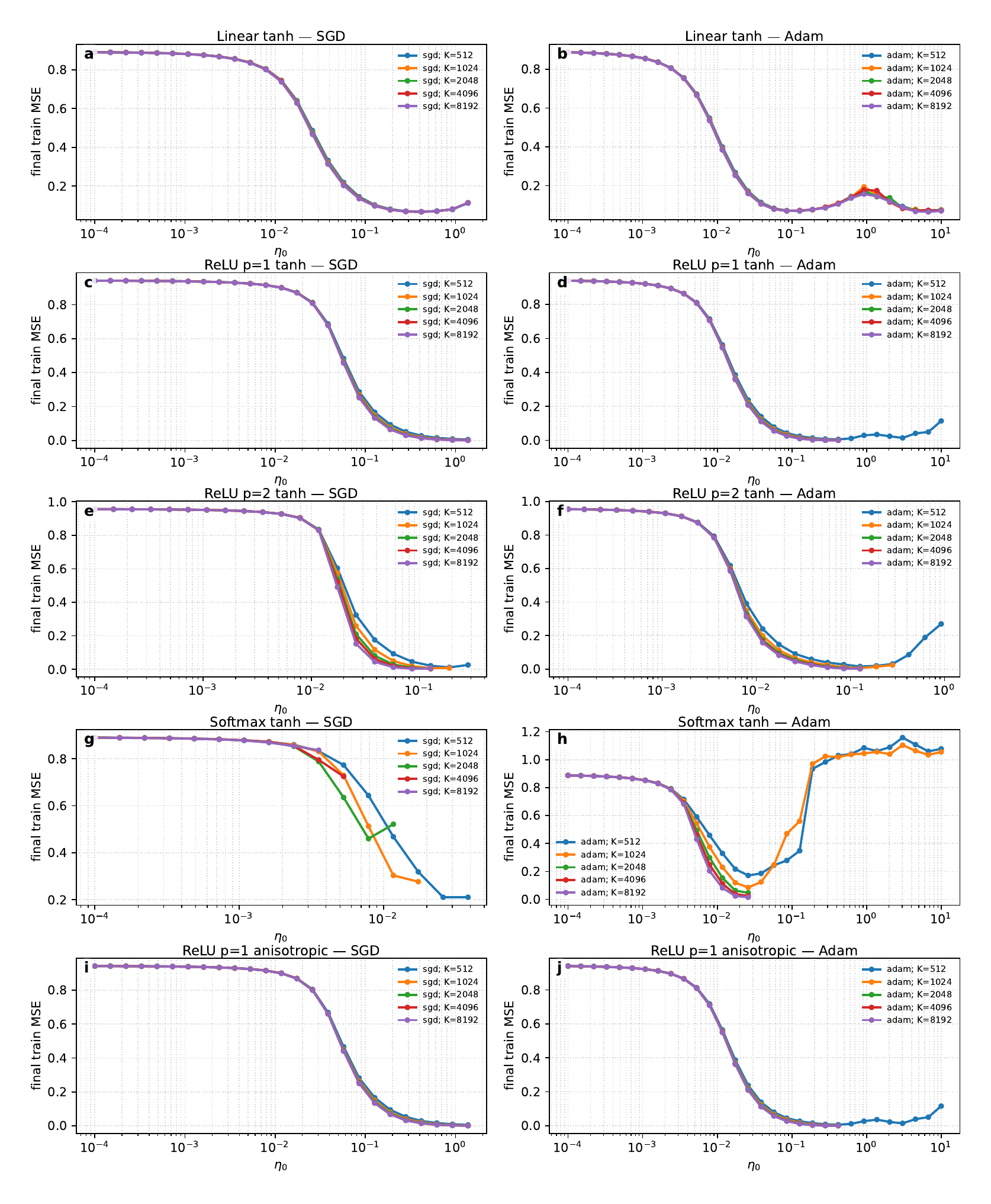}
    \caption{Learning rate transfer for SGD and Adam for width-only scaling of DenseAMs with ReLU and softmax activations, with $N=128, P=256, \beta=0.1$, trained for 256 epochs using the prescription in Table \ref{tab:results}. 
    For softmax with SGD, we used $s_1 = 1/ \sqrt{N}, s_2=\sqrt{N}, \eta_W = \eta_0 K$.
    Note the instability and lack of transfer for softmax DenseAMs trained with SGD. The anisotropic experiment shown in the bottom row was done with anisotropic inputs $x_\alpha \sim \mathcal N(0,D)$, where $D$ is a diagonal matrix with $i$-th entry proportional to $i^{-2/5}$ and trace $N$.} 
    \label{fig:transfter-K-only-regime}
\end{figure}

In the next sections we provide the scaling for Adam with elementwise activation as well as softmax.
See Fig. \ref{fig:transfter-K-only-regime} for HP transfer for linear, ReLU$^p$, and softmax DenseAM, for both SGD and Adam training.

\subsubsection{Adam updates of preactivations and outpus scaling}
\label{sec:adam-K-only-scaling}

For Adam, we use the same approximation in Eq. \eqref{eq:adam-sign-approx},
\begin{equation}
    U_{ki} := \Delta^{\rm Adam} \tilde W_{ki}\simeq -\eta_W \, \mathrm{sign}\left((\nabla_{\tilde W}\mcL_{\mathcal B})_{ki}\right).
\end{equation}
Thus each component of the update is \(U_{ki}=O_K(\eta_W)\). The preactivation update is
\begin{equation}
\label{eq:adam-dz}
    \Delta_{\rm Adam} Z_{k\mu}=s_1\sum_{i=1}^N U_{ki}G_{i\mu} \sim s_1 \eta_W \sqrt{N} \sim \eta_W,
\end{equation}
using $s_1 \sim 1/\sqrt{N}$ from \eqref{eq:K-only-s1-s2} (which does not scale with $K$).

The first-order output update is
\begin{equation}
    \Delta_{\rm Adam}F = s_2 U^\top \tilde S+s_2\tilde W^\top C\left(S'\odot \Delta_{\rm Adam}Z\right) :=  \Delta^{(1)}_{\rm Adam}F +  \Delta^{(2)}_{\rm Adam}F.
\end{equation}
For the first term,
\begin{equation}
\label{eq:adam-df1}
    \left( \Delta^{(1)}_{\rm Adam}F \right)_{i \mu} = \left(s_2 U^\top \tilde S\right)_{i\mu}=s_2\sum_{k=1}^K U_{ki}\tilde S_{k\mu}.
\end{equation}
In the only-$K$ regime, $N, B, P$ are fixed, so this is a sum over $K$ terms which are correlated as $U$ depends on the loss gradient \eqref{eq:loss-grad} which depends on $S$. 
Therefore, the sum generically adds up coherently, 
\begin{equation}
    \sum_{k=1}^K U_{ki}\tilde S_{k\mu} \sim \eta_W K.
\end{equation}
Plugging this estimate into \eqref{eq:adam-df1}, and using $s_2 \sim 1/K$ from \eqref{eq:K-only-s1-s2},
\begin{equation}
\label{eq:adam-df1-scale}
     \left( \Delta^{(1)}_{\rm Adam}F \right)_{i \mu} \sim s_2 \eta_W K \sim \eta_W.
\end{equation}
For the second term $\Delta^{(2)}_{\rm Adam}F$, we use \eqref{eq:adam-dz}, and again assuming the sum over $k$ adds up coherently, 
\begin{equation}
\label{eq:adam-df2-scale}
  \left( \Delta^{(2)}_{\rm Adam}F \right)_{i \mu} =  \left(s_2\tilde W^\top \left(S'\odot \Delta_{\rm Adam}Z\right) \right)_{i\mu} \sim s_2 K \eta_W \sim \eta_W
\end{equation}
Therefore both $\Delta_{\rm Adam}Z$, Eq. \eqref{eq:adam-dz} and $\Delta_{\rm Adam}F$, Eqs. \eqref{eq:adam-df1-scale}, \eqref{eq:adam-df2-scale} remain order one for
\begin{equation}
    s_1\sim 1/\sqrt N,\qquad s_2\sim 1/K,\qquad \eta_W\sim 1,\qquad \text{Adam, elementwise activations, \(K\)-only regime}.
\end{equation}
See Fig. \ref{fig:transfter-K-only-regime} for HP transfer using this prescription in the only-$K$ regime.

\subsection{Softmax activation}

In Sec. \ref{SI-sec:softmax-initialization} we analyzed the scales of the preactivations and the outputs. 
We have found that $s_1 \sim 1/\sqrt{N}$ (Eq. \eqref{SI-eq:softmax-s1}), and that $s_2 \sim 1 / s_1 \sim \sqrt{N}$ (see the discussion before Eq. \eqref{SI-eq:softmax-s2}).
Therefore in the only-$K$ regime we get
\begin{equation}
    s_1 \sim 1/\sqrt{N}, \quad s_2 \sim \sqrt{N},
\end{equation}
and both do not scale with $K$.

For the updates scaling we follow the arguments in Sec. \ref{SI-sec:softmax-adam-scaling}.
The update $\Delta z$ in \eqref{SI-eq:softmax-adam-dz-scaling} does not depend on $K$ or $s_2$, so it remains the same
\begin{equation}
    \Delta z_k \sim \eta_W.
\end{equation}
For the two output update terms $\Delta^{(1)}_{\rm softmax} f, \,  \Delta^{(2)}_{\rm softmax}$ in \eqref{SI-eq:softmax-adam-df-simplified}, we shall assume that the sums over $K$ terms in the only-$K$ regime (i.e. when $N, P, B$ are fixed) scale with $K$ instead of $\sqrt{K}$. 
This is similar to the behavior of the outputs in the simple linear case that was analyzed in Eq. \eqref{SI-eq:F-frob-final} where the difference between the proportional regime and the only-$K$ regime is explicit.
Therefore, Eq. \eqref{SI-eq:softmax-adam-df-term1-Keff} becomes
\begin{equation}
    \left( \Delta^{(1)}_{\rm softmax} f\right)_i = s_2 \sum_{k=1}^K U_{ki} (C \sigma)_k  \sim s_2  \frac{\eta_W}{K} \sum_{k=1}^K \xi_{ki} \sim s_2 \eta_W \sim \eta_W,
\end{equation}
where in the first transition we used the definition, then, we pulled out the scales $U_{ki} \sim \eta_W, \, (C\sigma)_k \sim 1/K$ and introduced the order one $\xi_{ki}$ terms.
Then, we assumed the sum adds up coherently in the large $K$ limit (where $N, P, B$ are fixed), and finally, we used $s_2 \sim \sqrt{N} \sim 1$.

For the term $\Delta^{(2)}_{\rm softmax} f$,  Eq. \eqref{SI-eq:softmax-adam-df-term2} is
\begin{equation}
    \left( \Delta^{(2)}_{\rm softmax} f\right)_i = s_2 \sum_{k=1}^K W_{ki} (J \Delta Z)_k  .
\end{equation}
We use $(J \Delta Z)_k \sim \xi_k \eta_W / K, \, \xi_k = O(1)$, from \eqref{SI-eq:softmax-adam-Jdz-scaling}, and the above becomes
\begin{equation}
    \left( \Delta^{(2)}_{\rm softmax} f\right)_i = s_2 \sum_{k=1}^K W_{ki} (J \Delta Z)_k \sim s_2 \frac{\eta_W}{K} \sum_{k=1}^K W_{ki} \xi_k \sim s_2 \eta_W \sim \eta_W ,
\end{equation}
where in the third transition we again assumed a coherent sum $\sum_{k=1}^K W_{ki} \xi_k \sim K$, and lastly, we used $s_2 \sim 1$.

Finally, we arrive at the only-$K$ prescription for Adam,
\begin{equation}
    s_1\sim 1/\sqrt N,\qquad s_2\sim \sqrt{N},\qquad \eta_W\sim 1,\qquad \text{Adam, softmax, \(K\)-only regime}.
\end{equation}
See Fig. \ref{fig:transfter-K-only-regime} for HP transfer using this prescription in the only-$K$ regime.

\section{Experiments with MNIST}
\label{sec:mnist}

In this section we report empirics about HP transfer and consistency across scale for DenseAMs \eqref{eq:f-def} trained on MNIST in the proportional regime \eqref{eq:proportional-regime}. While MNIST is not a particularly interesting dataset, we use it showcase two points: (a) that we get transfer even on highly non-isotropic non-Gaussian data (b) scaling the input dimension, together with model size, is possible even for a fixed dataset. 

\begin{figure}
    \centering
    \includegraphics[width=1\linewidth]{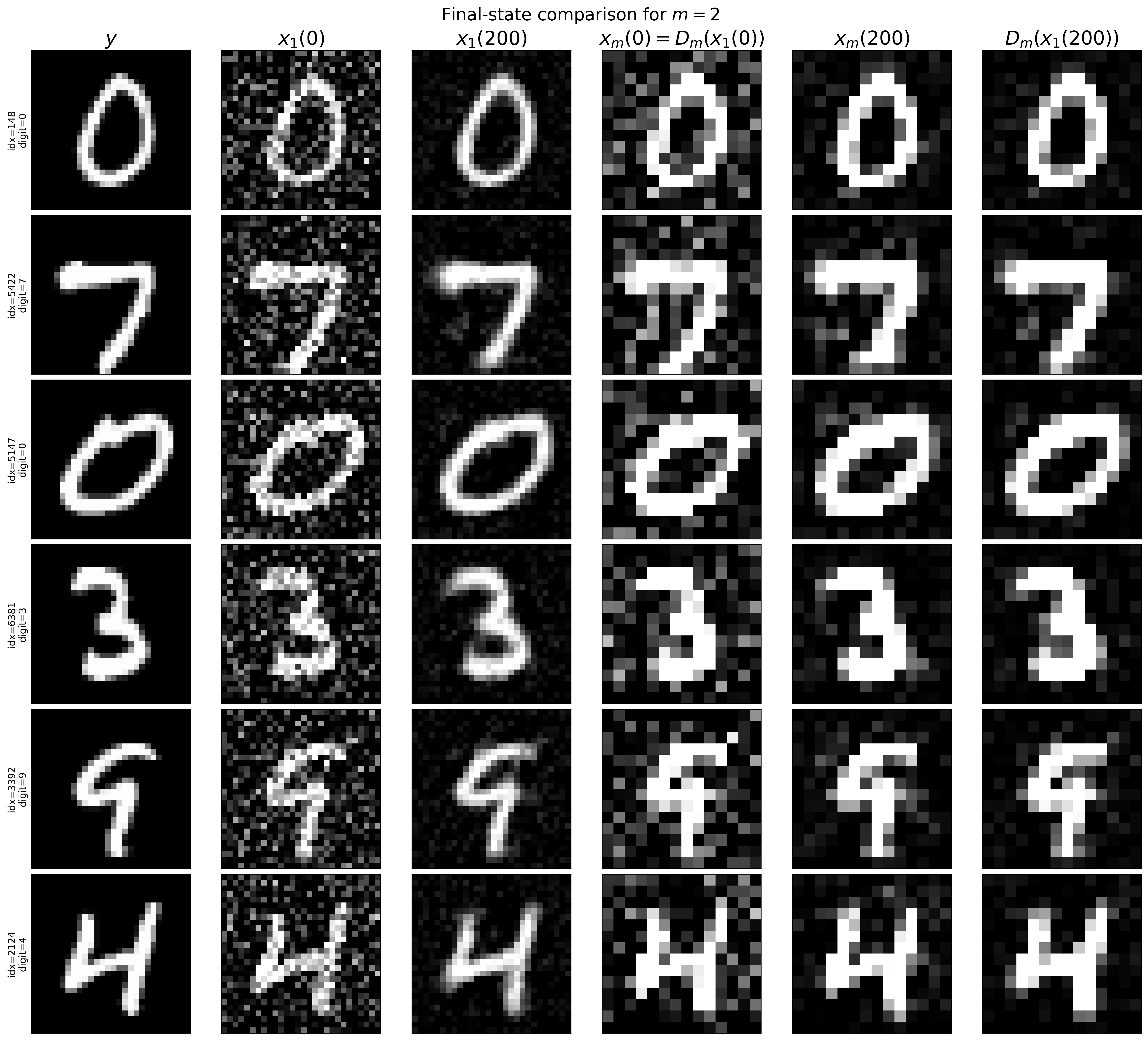}
    \caption{Comparison of denoisers across $N,K,P,B$ for DenseAMs trained on MNIST digits. The left three columns are the unnoised image $y$, the corrupted image $x_1(0)$ and the denoised output $x_1(200)$ for a DenseAM trained with input dimension $N=784$. The final column records the factor of $m=2$ plaquette downsampling $D_m(x_1(200))$ of $x_1(200)$. The fourth column is factor of $m$  downsampling $D_m(x_1(0))$ of the original corrupted image and the fifth column is the denoising of that downsampled version of $x_1(0)$, i.e., $D_{m=2}(x_1(200))$, by a factor of $m$ smaller DenseAM which is trained on a factor of $m$ downsampled MNIST data.
    Comparing the fifth column (the output of the downscaled DenseAM) with the sixth column (the downscled output of the original full-size DenseAM) exhibits good agreement.}
    \label{fig:mnist-examples}
\end{figure}

We generate versions of the MNIST images with input dimension $N\leq N_0=784$ by coarse graining the original \(28\times 28\) pixel grid. Specifically, for a chosen block size \(j\), we define a linear downscaling map
\[
D_j:\mathbb{R}^{784}\to\mathbb{R}^{N_j},
\]
and replace each image \(x_{1}\) (where \(x_{1}\) denotes an original image) by its coarse grained version
\[
x_{j} = D_j x_{1}.
\]
While we explored several types of downscaling we found that  the plaquette reduction works best. This means we average over \(j\times j\) blocks of neighboring pixels reducing both row and columnn dimension simultaneously and obtain
\[
N_j=\Bigl\lceil \frac{28}{j}\Bigr\rceil \times \Bigl\lceil \frac{28}{j}\Bigr\rceil.
\]
The same coarse graining is applied to the network weights.
If \(w_\alpha\in\mathbb{R}^{784}\) denotes the \(\alpha\)-th row of the weight matrix in the original space, then its reduced version is obtained by projection onto the coarse variables,
\[
w_\alpha^{(j)} = w_\alpha D_j^{\top},
\]
so that the row is transformed consistently with the reduced representation of the data. Before training the DenseAM, and after downscaling, we center each pixel of the image across the entire dataset. Our main findings are as follows:
\begin{itemize}
    \item \textbf{HP Transfer.} The plaquette scaling described above yields HP transfer across $N,K,P,B$ in the proportional regime \eqref{eq:proportional-regime}. See Figure \ref{fig:transfer-mnist}.
    \item \textbf{Consistent Denoising.} In figure \ref{fig:mnist-examples} we compare training a large DenseAM on MNIST, performing $200$ steps of the denoising on a fixed input, and then downsampling the result with first training a smaller DenseAM on downsampled MNIST and then denoising a downsampled version of the same input.
\end{itemize}

\section{Information about numerical experiments}
\label{sec:experiments-info}

All numerical experiments were performed on a standard Apple MacBook Pro M4 laptop and take order of minutes to run.
Details of the experiments are provided in the figure captions where they are shown.
For all experiments we used random seeds at each run. Curves are always reproduced.
When training with Adam, for all experiments, we set in Eqs. \eqref{SI-eq:adam-mv}, \eqref{SI-eq:adam-update}, $\beta_1 = 0.9, \beta_2=0.999, \epsilon_{\rm Adam}=10^{-8}$.

\end{document}